\documentclass[twoside,letterpaper]{article}

\usepackage[preprint]{aistats2027}
\usepackage[T1]{fontenc}

\usepackage[whole,ipaex-type1]{bxcjkjatype}

\usepackage{amsmath}   
\usepackage{amsfonts}  
\usepackage{amssymb}   
\usepackage{amsthm}    
\usepackage{mathtools} 
\usepackage{bm}        

\usepackage{graphicx}
\usepackage[caption=false]{subfig}
\usepackage{placeins} 

\usepackage{booktabs}
\usepackage{multirow}

\usepackage{algorithm}
\usepackage[noend]{algpseudocode} 

\theoremstyle{definition}
\newtheorem{definition}{Definition}[section]
\theoremstyle{plain}
\newtheorem{proposition}{Proposition}[section]

\usepackage[round]{natbib}

\usepackage[hidelinks,hypertexnames=false]{hyperref}
\usepackage[capitalise,noabbrev,nameinlink]{cleveref}
\crefname{equation}{Eq.}{Eqs.}
\Crefname{equation}{Eq.}{Eqs.}
\crefname{proposition}{Proposition}{Propositions}
\Crefname{proposition}{Proposition}{Propositions}

\crefalias{ALG@line}{line}
\crefname{line}{Line}{Lines}
\Crefname{line}{Line}{Lines}
\crefrangeformat{line}{#3Lines~#1#4--#5#2#6}
\Crefrangeformat{line}{#3Lines~#1#4--#5#2#6}
\makeatletter
\AddToHook{env/algorithmic/begin}{\def\@currentcounter{ALG@line}}
\makeatother

\usepackage{autonum}

\begin{document}

\runningauthor{Akira Ito, Daiki Chijiwa, Atsutoshi Kumagai}

\twocolumn[
\aistatstitle{Permutation-Based Pre-trained Model Merging via Reciprocal Orthogonality}
\aistatsauthor{Akira Ito \And Daiki Chijiwa \And Atsutoshi Kumagai}
\aistatsaddress{Tohoku University
\And NTT Computer and \\ Data Science Laboratories
\And NTT Computer and \\ Data Science Laboratories}
]

\begin{abstract}
 \citet{Ito_ICLR_2026} showed that sufficiently wide, independently trained neural networks can exhibit linear mode connectivity (LMC) with softmax temperature calibration, even without permutation. LMC means that a low-loss linear path connects the models' parameters. This behavior is explained by layerwise exponentially weighted connectivity (LEWC), whereby each merged layer outputs an exponentially weighted sum of the original models' corresponding representations, yielding ensemble-like predictions. However, prior work observed LEWC with substantial widening, such as 8192 units per hidden layer for MLPs and $16\times$ width for VGG-11, raising the question of whether such widths are necessary. We therefore search for permutations that encourage reciprocal orthogonality (RO), which suppresses interference between each model's weights and the other model's representations. This allows both models' computations to coexist in the merged network, promoting LEWC at smaller widths. We formulate the search as independent layerwise quadratic assignment problems and approximately solve their continuous relaxations using Frank--Wolfe. Experiments show that our method facilitates LMC with less width expansion. It also merges models trained on different data distributions, where weight matching, which maximizes weight similarity through permutations, struggles. The merged predictions approximate those of a logit ensemble of the original models.
\end{abstract}

\section{Introduction}
\label{sec:introduction}

Large neural networks are widely used across diverse domains \citep{Krizhevsky_NIPS_2012,Aaron_arxiv_2016,Vaswani_NIPS_2017}.
Although their training involves large-scale nonconvex optimization, simple methods such as stochastic gradient descent (SGD) often find solutions with good predictive performance.
Independent runs with different initializations and mini-batch orders yield distinct parameter values but similar performance, raising the question of why optimization works so well.

One possible explanation is that loss landscapes have a simpler structure than expected.
Previous studies have empirically shown that independently trained solutions can be connected by simple nonlinear low-loss paths, a property known as mode connectivity \citep{Garipov_NIPS_2018,Draxler_ICML_2018}.

Furthermore, \citet{Entezari_arxiv_2022} conjectured that accounting for permutation symmetries allows independently trained solutions to be mapped into a single approximately convex low-loss basin.
These symmetries arise from parameter redundancies that allow hidden units and the corresponding weights in adjacent layers to be reordered without changing a network's input--output mapping.
The approximate convexity of this basin means that, after appropriate permutations, independently trained models can be connected by low-loss linear paths, a property known as linear mode connectivity (LMC) \citep{frankle_ICML_2020}.

\citet{Ainsworth_ICLR_2023} provided empirical support for this picture in MLPs, VGG, and ResNet through methods including weight matching (WM), which searches for permutations that reduce the distance between model weights, thereby increasing parameter similarity and promoting LMC.
Prior studies~\citep{Zhou_NIPS_2023,Ito_ICLR_2025} have analytically demonstrated that LMC in conventional permutation-based methods, with WM as a representative example, is achieved by increasing similarity between the two models' hidden-layer functions, which helps preserve performance under parameter interpolation.

More recently, \citet{Ito_ICLR_2026} showed that sufficiently wide, independently trained networks can lie in a common low-loss basin under softmax temperature calibration even without permutations.
They explained this behavior by introducing layerwise exponentially weighted connectivity (LEWC), whereby interpolated layers output exponentially weighted sums of the original models' corresponding representations.
Reciprocal orthogonality (RO), one of the jointly sufficient conditions for LEWC, suppresses interference between each model's weights and the other's features, allowing their computations to proceed approximately independently within the interpolated network.
The resulting logits are proportional to those of a weighted logit ensemble, with temperature calibration correcting their scale.

In this work, we investigate whether accounting for permutation symmetries can reveal LEWC-based LMC under temperature calibration without the extensive widening used by \citet{Ito_ICLR_2026}, such as 8192 units per hidden layer for MLPs and $16\times$ width for VGG-11.
Our goal is to understand how network width and permutation symmetries jointly shape the connectivity of independently trained solutions.
Clarifying this relationship contributes to our understanding of why neural networks can be trained effectively despite their nonconvex loss landscapes.
To this end, we search for permutations that promote RO and suppress interference between models.
Our contributions are threefold.

\begin{enumerate}
\item \textbf{Permutation search to promote LEWC.}
We formulate permutation search to promote RO as independent layerwise quadratic assignment problems (QAPs).
Following the fast approximate QAP (FAQ) approach \citep{Vogelstein_PLOSONE_2015}, we apply the Frank--Wolfe algorithm to a relaxation over doubly stochastic matrices and discretize the solutions into permutations.
We propose two variants, using hidden-layer activations or weights alone.
Experiments show that our method promotes LEWC and facilitates LMC under temperature calibration with less width expansion than permutation-free interpolation \citep{Ito_ICLR_2026}.

\item \textbf{Characterizing the low-loss region under temperature calibration.}
We empirically show that, after applying the discovered permutations, calibrated loss remains low throughout the triangle spanned by the two models and their parameter sum.
This extends the picture of connectivity beyond the linear path between the original models to a low-loss region containing both models and their sum.

\item \textbf{Merging models across training distributions by parameter addition.}
We apply our method to model merging, which combines the parameters of trained models into a single network.
Under the conditions for LEWC, parameter addition yields the sum of the original models' logits, avoiding the decrease in activation magnitude with depth caused by parameter averaging.
Motivated by this property, we merge models by adding their parameters after permutation.
This approach succeeds in settings where WM struggles, including MLPs trained separately on MNIST and Fashion-MNIST, and classifiers trained on disjoint five-class subsets of CIFAR-10.
The merged predictions approximate those of logit ensembles of the original models, supporting the use of LEWC to integrate distinct predictive functions into one network.
\end{enumerate}

\section{Preliminaries}
\label{sec:preliminaries}

\subsection{Notation}\label{sec:notation}

For a positive integer $k$, let $[k]=\{1,2,\ldots,k\}$. The Frobenius norm of a matrix $\bm{X}$ is $\|\bm{X}\|_F=(\sum_{i,j}X_{ij}^2)^{1/2}$, and the Euclidean norm of a vector $\bm{x}$ is $\|\bm{x}\|_2$. For matrices $\bm{A}$ and $\bm{B}$ of the same shape, define the Frobenius inner product as $\langle\bm{A},\bm{B}\rangle_F=\operatorname{tr}(\bm{A}^\top\bm{B})=\sum_{i,j}A_{ij}B_{ij}$. Let $\bm{I}_d$ denote the $d$-dimensional identity matrix, $\ker\bm{X}$ the null space of $\bm{X}$, and $\mathcal{P}_d$ the set of $d\times d$ permutation matrices.

For simplicity, we consider a multilayer perceptron (MLP) with $L$ fully connected layers. The same basic ideas apply to other architectures. Let $d_\ell$, $\bm{W}_\ell\in\mathbb{R}^{d_\ell\times d_{\ell-1}}$, and $\bm{b}_\ell\in\mathbb{R}^{d_\ell}$ be the width, weight matrix, and bias vector of layer $\ell$, respectively, and let $\bm{\theta}$ collect all vectorized parameters. For an input $\bm{x}\in\mathbb{R}^{d_0}$, set $\bm{a}_0=\bm{x}$ and define the pre-activation and activation at each hidden layer $\ell\in[L-1]$ as $\bm{z}_\ell=\bm{W}_\ell\bm{a}_{\ell-1}+\bm{b}_\ell$ and $\bm{a}_\ell=\sigma(\bm{z}_\ell)$, where $\sigma$ is the ReLU function. The final layer is linear with classification logits $\bm{z}_L=\bm{W}_L\bm{a}_{L-1}+\bm{b}_L$. For notational consistency across layers, we define $\bm{a}_L=\bm{z}_L$ and write $f(\bm{x};\bm{\theta})=\bm{a}_L$.

\subsection{Permutation Invariance}\label{sec:permutation-invariance}

In an MLP, reordering hidden units and transforming the adjacent weight matrices accordingly leaves the network's input--output mapping unchanged. Let $\bm{P}_\ell\in\mathcal{P}_{d_\ell}$ be a permutation matrix for hidden layer $\ell\in[L-1]$. Since $\bm{P}_\ell^\top\bm{P}_\ell=\bm{I}_{d_\ell}$ and permutations commute with elementwise activation functions, we have
$\bm{a}_\ell
=\bm{P}_\ell^\top\bm{P}_\ell\bm{a}_\ell
=\bm{P}_\ell^\top\sigma\!\left(
\bm{P}_\ell\bm{W}_\ell\bm{a}_{\ell-1}
+\bm{P}_\ell\bm{b}_\ell\right)$. Thus, $\bm{P}_\ell$ can be absorbed into the weights and biases of layer $\ell$, and $\bm{P}_\ell^\top$ into the next layer's weights. To permute all hidden layers simultaneously, set $\bm{P}_0=\bm{I}_{d_0}$ and $\bm{P}_L=\bm{I}_{d_L}$, and transform each layer $\ell\in[L]$ as $\widetilde{\bm{W}}_\ell=\bm{P}_\ell\bm{W}_\ell\bm{P}_{\ell-1}^\top$ and $\widetilde{\bm{b}}_\ell=\bm{P}_\ell\bm{b}_\ell$. We denote this transformation by $\pi$ and use tildes for the permuted quantities. The hidden activations satisfy $\widetilde{\bm{a}}_\ell=\bm{P}_\ell\bm{a}_\ell$, while the output coordinates remain fixed. Hence, $f(\bm{x};\pi(\bm{\theta}))=f(\bm{x};\bm{\theta})$ for every input. This permutation symmetry changes the relationship between models in parameter space while preserving each model's predictions \citep{Entezari_arxiv_2022,Ainsworth_ICLR_2023}.

\subsection{LMC through RO}\label{sec:ro-lmc}

We first define LMC and LEWC, then introduce two conditions that jointly imply LEWC and explain how LEWC relates to LMC through temperature calibration.
\begin{definition}[LMC, {\citealp{frankle_ICML_2020,Entezari_arxiv_2022}}]
\label{def:lmc}
Let $\mathcal{L}(\bm{\theta})$ be the model's loss on a common evaluation distribution. For two models $\bm{\theta}^{(a)}$ and $\bm{\theta}^{(b)}$, define their linear interpolation as $\bm{\theta}^{(c)}=\lambda\bm{\theta}^{(a)}+(1-\lambda)\bm{\theta}^{(b)}$ for $\lambda\in[0,1]$, and their loss barrier as
$B(\bm{\theta}^{(a)},\bm{\theta}^{(b)})
=\max_{\lambda\in[0,1]}\bigl(
\mathcal{L}(\bm{\theta}^{(c)})
-\lambda\mathcal{L}(\bm{\theta}^{(a)})
-(1-\lambda)\mathcal{L}(\bm{\theta}^{(b)})\bigr)$. The two models exhibit LMC when this barrier is sufficiently small.

Following \citet{Ito_ICLR_2026}, we also consider LMC under temperature calibration. This property holds when applying softmax temperature calibration to every model along the linear path yields a sufficiently small barrier in calibrated loss.
\end{definition}

\textbf{LEWC.} To explain LMC under temperature calibration in wide networks, \citet{Ito_ICLR_2026} introduced LEWC, a property under which each layer of the merged model produces a weighted sum of the corresponding layer outputs in the original models.
\begin{definition}[LEWC, {\citealp{Ito_ICLR_2026}}]
\label{def:lewc}
Two models satisfy layerwise exponentially weighted connectivity (LEWC) if, for every $\lambda\in[0,1]$, the merged model $\bm{\theta}^{(c)}=\lambda\bm{\theta}^{(a)}+(1-\lambda)\bm{\theta}^{(b)}$ satisfies $\bm{a}_\ell^{(c)}=\lambda^\ell\bm{a}_\ell^{(a)}+(1-\lambda)^\ell\bm{a}_\ell^{(b)}$ for the inputs under consideration and every layer $\ell\in[L]$.
\end{definition}

\textbf{Two sufficient conditions for LEWC.} We next introduce ReLU weak additivity and RO, which together provide sufficient conditions for LEWC. Following \citet{Ito_ICLR_2026}, we omit biases in the following analysis for simplicity.

\begin{definition}[ReLU weak additivity, {\citealp{Zhou_NIPS_2023,Ito_ICLR_2026}}]
\label{def:weak-additivity}
Weak additivity for the ReLU activation function holds if, for every hidden layer $\ell\in[L-1]$ and every $\lambda\in[0,1]$, the original models' pre-activations $\bm{z}_\ell^{(a)}$ and $\bm{z}_\ell^{(b)}$ satisfy
\begin{align}
\sigma\!\left(\lambda\bm{z}_\ell^{(a)}+(1-\lambda)\bm{z}_\ell^{(b)}\right)
=\lambda\sigma(\bm{z}_\ell^{(a)})+(1-\lambda)\sigma(\bm{z}_\ell^{(b)}).
\end{align}
\end{definition}
This condition requires ReLU to act linearly between the two pre-activation vectors under consideration. Although it does not hold for arbitrary inputs, prior work has observed it empirically \citep{Ito_ICLR_2026,Zhou_NIPS_2023}.

\begin{definition}[Reciprocal orthogonality, {\citealp{Ito_ICLR_2026}}]
\label{def:reciprocal-orthogonality}
Two models satisfy RO if $\bm{W}_{\ell+1}^{(a)}\bm{a}_\ell^{(b)}=\bm{0}$ and $\bm{W}_{\ell+1}^{(b)}\bm{a}_\ell^{(a)}=\bm{0}$ for every hidden layer $\ell\in[L-1]$.
\end{definition}
This means that one model's representation at layer $\ell$ lies in the null space of the other model's weight matrix at layer $\ell+1$, eliminating cross terms during merging. It is distinct from orthogonality between the two models' parameter vectors.

Weak additivity and RO imply LEWC, as can be shown by induction on the layer index \citep{Ito_ICLR_2026}. In their experiments, \citet{Ito_ICLR_2026} found that substantial widening was needed for these two conditions, and consequently LEWC, to hold.

\textbf{Activation scaling and its relation to LMC.} Under LEWC, equal-weight merging ($\lambda=1/2$) gives $\bm{a}_\ell^{(c)}=2^{-\ell}(\bm{a}_\ell^{(a)}+\bm{a}_\ell^{(b)})$, which scales the mean of the original activations by a factor of $2^{1-\ell}$. For general $\lambda$, the logits coincide, up to a positive scale factor, with a logit ensemble whose weights are in the ratio $\lambda^L:(1-\lambda)^L$. The merged model therefore predicts the same labels as this ensemble and achieves high accuracy whenever the ensemble does. However, cross-entropy depends on logit scale, so a reduction in this scale can produce a loss barrier. Softmax temperature calibration adjusts this scale without changing the predicted labels. \citet{Ito_ICLR_2026} showed empirically that this calibration yields low-loss linear paths, establishing LMC under temperature calibration.

\section{Proposed Method}
\label{sec:proposed_method}

We propose a permutation search method that promotes RO between two given models, enabling model merging through LEWC without extensive width expansion. Among the sufficient conditions for LEWC, ReLU weak additivity has been reported to hold approximately in many cases \citep{Ito_ICLR_2026,Zhou_NIPS_2023}. We therefore focus on promoting the other condition, RO, through parameter permutations. As in \cref{sec:ro-lmc}, we omit biases below.

In \cref{sec:ro-activation,sec:ro-weight}, we formulate independent layerwise quadratic assignment problems (QAPs) using activations or weights alone. In \cref{sec:ro-frank-wolfe}, we describe how to solve their continuous relaxations using Frank--Wolfe.

\subsection{Formulating RO Using Activations}
\label{sec:ro-activation}

We first formulate permutation search for RO using activations. Consider two models $\bm{\theta}^{(a)}$ and $\bm{\theta}^{(b)}$ with the same architecture, and apply a permutation $\pi$ to the latter. By \cref{sec:permutation-invariance}, the permuted weights and activations satisfy $\widetilde{\bm{W}}_{\ell+1}^{(b)}=\bm{P}_{\ell+1}\bm{W}_{\ell+1}^{(b)}\bm{P}_\ell^\top$ and $\widetilde{\bm{a}}_\ell^{(b)}=\bm{P}_\ell\bm{a}_\ell^{(b)}$. To optimize $\bm{P}_\ell$ at hidden layer $\ell\in[L-1]$, we focus on the activations at layer $\ell$ and weights at layer $\ell+1$.

For $\bm{\theta}^{(a)}$ and $\pi(\bm{\theta}^{(b)})$ to satisfy RO at this hidden layer, \cref{def:reciprocal-orthogonality} requires both $\bm{W}_{\ell+1}^{(a)}\widetilde{\bm{a}}_\ell^{(b)}=\bm{0}$ and $\widetilde{\bm{W}}_{\ell+1}^{(b)}\bm{a}_\ell^{(a)}=\bm{0}$. Substituting the permuted weights and activations yields two conditions $\bm{W}_{\ell+1}^{(a)}\widetilde{\bm{a}}_\ell^{(b)}
=\bm{W}_{\ell+1}^{(a)}\bm{P}_\ell\bm{a}_\ell^{(b)}=\bm{0}$ and $
\widetilde{\bm{W}}_{\ell+1}^{(b)}\bm{a}_\ell^{(a)}
=\bm{P}_{\ell+1}\bm{W}_{\ell+1}^{(b)}\bm{P}_\ell^\top\bm{a}_\ell^{(a)}=\bm{0}$.
Each condition is equivalent to the squared norm of its left-hand side being zero. Since permutation matrices preserve the Euclidean norm, the second condition gives $\|\bm{P}_{\ell+1}\bm{W}_{\ell+1}^{(b)}\bm{P}_\ell^\top\bm{a}_\ell^{(a)}\|_2^2=\|\bm{W}_{\ell+1}^{(b)}\bm{P}_\ell^\top\bm{a}_\ell^{(a)}\|_2^2$, allowing us to remove the output permutation $\bm{P}_{\ell+1}$. We thus define the sum of the two squared norms as
$r_\ell(\bm{P}_\ell;\bm{x})
=\|\bm{W}_{\ell+1}^{(a)}\bm{P}_\ell\bm{a}_\ell^{(b)}\|_2^2
+\|\bm{W}_{\ell+1}^{(b)}\bm{P}_\ell^\top\bm{a}_\ell^{(a)}\|_2^2$.
The activations $\bm{a}_\ell^{(a)}$ and $\bm{a}_\ell^{(b)}$ depend on the input $\bm{x}$. Since both terms are nonnegative, satisfying the RO conditions at this hidden layer almost surely over $\bm{x}$ is equivalent to $r_\ell(\bm{P}_\ell;\bm{x})=0$ almost surely.

Let $\mathcal{D}$ be the input distribution of interest. We minimize $J_\ell^{\mathrm{act}}(\bm{P}_\ell)=\mathbb{E}_{\bm{x}\sim\mathcal{D}}[r_\ell(\bm{P}_\ell;\bm{x})]$. For $m\in\{a,b\}$, let $\bm{G}_{\mathrm{col},\ell+1}^{(m)}=(\bm{W}_{\ell+1}^{(m)})^\top\bm{W}_{\ell+1}^{(m)}$ be the Gram matrix of the weight columns, and let $\bm{C}_\ell^{(m)}=\mathbb{E}_{\bm{x}\sim\mathcal{D}}[\bm{a}_\ell^{(m)}(\bm{a}_\ell^{(m)})^\top]$ be the second-moment matrix of the activations. Using $\|\bm{v}\|_2^2=\operatorname{tr}(\bm{v}\bm{v}^\top)$ and the cyclic property of the trace, the expectation of $r_\ell(\bm{P}_\ell;\bm{x})$ becomes
\begin{align}
J_\ell^{\mathrm{act}}(\bm{P}_\ell)
&=\mathbb{E}_{\bm{x}\sim\mathcal{D}}\bigl[
\operatorname{tr}\bigl(\bm{G}_{\mathrm{col},\ell+1}^{(a)}\bm{P}_\ell
\bm{a}_\ell^{(b)}(\bm{a}_\ell^{(b)})^\top\bm{P}_\ell^\top\bigr)
\nonumber\\
&\hphantom{{}=\mathbb{E}_{\bm{x}\sim\mathcal{D}}\bigl[}{}+\operatorname{tr}\bigl(\bm{G}_{\mathrm{col},\ell+1}^{(b)}\bm{P}_\ell^\top
\bm{a}_\ell^{(a)}(\bm{a}_\ell^{(a)})^\top\bm{P}_\ell\bigr)\bigr]
\nonumber\\
&=\operatorname{tr}\bigl(\bm{G}_{\mathrm{col},\ell+1}^{(a)}\bm{P}_\ell\bm{C}_\ell^{(b)}\bm{P}_\ell^\top\bigr)
\nonumber\\
&\quad{}\quad\quad\quad+\operatorname{tr}\bigl(\bm{G}_{\mathrm{col},\ell+1}^{(b)}\bm{P}_\ell^\top\bm{C}_\ell^{(a)}\bm{P}_\ell\bigr).
\label{eq:ro-activation-objective}
\end{align}
If this value is zero, the RO conditions at this hidden layer hold almost everywhere with respect to $\mathcal{D}$. Since a permutation attaining zero need not exist, we instead consider the minimization problem $\min_{\bm{P}\in\mathcal{P}_{d_\ell}}J_\ell^{\mathrm{act}}(\bm{P})$. As \cref{eq:ro-activation-objective} is quadratic in the entries of the permutation matrix, this is a QAP. We solve it approximately through the continuous relaxation and discretization described in \cref{sec:ro-frank-wolfe}. The objective does not depend on permutations at other layers, so each $\bm{P}_\ell$ can be optimized independently. We estimate $\bm{C}_\ell^{(m)}$ by averaging outer products of activations obtained from unlabeled input samples.

\subsection{Approximating RO Using Only Weights}
\label{sec:ro-weight}

We next develop a formulation that searches for permutations promoting RO using only model parameters, without requiring activations or input data. We replace ReLU with the identity and approximate $\bm{a}_\ell^{(m)}\approx\bm{W}_\ell^{(m)}\bm{a}_{\ell-1}^{(m)}$. Our experiments in \cref{sec:exp-same-conditions} support this approximation, showing little or no reduction in the performance of merged models compared with the activation-based formulation in most settings. Under this linearization, the conditions
\begin{align}
\bm{W}_{\ell+1}^{(a)}\bm{P}_\ell\bm{W}_\ell^{(b)}&=\bm{0},
&\bm{W}_{\ell+1}^{(b)}\bm{P}_\ell^\top\bm{W}_\ell^{(a)}&=\bm{0}
\label{eq:ro-weight-condition}
\end{align}
eliminate both cross terms regardless of $\bm{a}_{\ell-1}^{(a)}$ and $\bm{a}_{\ell-1}^{(b)}$. Thus, \cref{eq:ro-weight-condition} is sufficient for RO in the linearized models.

These conditions lead to an objective computable from weights alone. Let $\bm{G}_{\mathrm{row},\ell}^{(m)}=\bm{W}_\ell^{(m)}(\bm{W}_\ell^{(m)})^\top$ be the Gram matrix of the weight rows. Define
\begin{align}
J_\ell^{\mathrm{weight}}(\bm{P}_\ell)
&=\|\bm{W}_{\ell+1}^{(a)}\bm{P}_\ell\bm{W}_\ell^{(b)}\|_F^2
+\|\bm{W}_{\ell+1}^{(b)}\bm{P}_\ell^\top\bm{W}_\ell^{(a)}\|_F^2
\nonumber\\
&=\operatorname{tr}\bigl(\bm{G}_{\mathrm{col},\ell+1}^{(a)}\bm{P}_\ell\bm{G}_{\mathrm{row},\ell}^{(b)}\bm{P}_\ell^\top\bigr)
\nonumber\\
&\quad{}\quad\quad+\operatorname{tr}\bigl(\bm{G}_{\mathrm{col},\ell+1}^{(b)}\bm{P}_\ell^\top\bm{G}_{\mathrm{row},\ell}^{(a)}\bm{P}_\ell\bigr).
\label{eq:ro-weight-objective}
\end{align}
Minimizing \cref{eq:ro-weight-objective} over $\bm{P}_\ell\in\mathcal{P}_{d_\ell}$ also gives an independent QAP for each layer. This objective replaces $\bm{C}_\ell^{(m)}$ in \cref{eq:ro-activation-objective} with $\bm{G}_{\mathrm{row},\ell}^{(m)}$ and requires neither activations nor input data. It serves as a surrogate that promotes RO, rather than a guarantee of exact RO in the original ReLU networks.

\subsection{Permutation Search via Frank--Wolfe}
\label{sec:ro-frank-wolfe}

Following the fast approximate QAP (FAQ) approach \citep{Vogelstein_PLOSONE_2015}, we approximately solve the QAPs defined by \cref{eq:ro-activation-objective,eq:ro-weight-objective} by applying Frank--Wolfe to their continuous relaxations. For a fixed layer $\ell$, let $J_\ell$ denote either $J_\ell^{\mathrm{act}}$ or $J_\ell^{\mathrm{weight}}$.

We relax the search space from the set of permutation matrices $\mathcal{P}_{d_\ell}$ to its convex hull, the set of doubly stochastic matrices $\mathcal{B}_{d_\ell}=\{\bm{X}\in\mathbb{R}^{d_\ell\times d_\ell}:\bm{X}\geq\bm{0},\ \bm{X}\bm{1}=\bm{1},\ \bm{X}^\top\bm{1}=\bm{1}\}$. Here, $\bm{1}$ is the all-ones vector, and the inequality is elementwise. Set $\bm{S}_\ell^{(m)}=\bm{C}_\ell^{(m)}$ for the activation formulation and $\bm{S}_\ell^{(m)}=\bm{G}_{\mathrm{row},\ell}^{(m)}$ for the formulation using only weights. The relaxed objective shared by both methods is
\begin{align}
J_\ell(\bm{X})
&=\operatorname{tr}\bigl(\bm{G}_{\mathrm{col},\ell+1}^{(a)}\bm{X}\bm{S}_\ell^{(b)}\bm{X}^\top\bigr)
\nonumber\\
&\quad{}+\operatorname{tr}\bigl(\bm{G}_{\mathrm{col},\ell+1}^{(b)}\bm{X}^\top\bm{S}_\ell^{(a)}\bm{X}\bigr),
\label{eq:ro-relaxed-objective}
\end{align}
which we minimize over $\bm{X}\in\mathcal{B}_{d_\ell}$.

The basic idea is to use the gradient to select a permutation that provides a promising update direction, then move toward it to decrease the objective. At iteration $i$, the first-order approximation around the current iterate $\bm{X}_i$ estimates the objective at a permutation $\bm{P}$ as
$J_\ell(\bm{P})
\approx J_\ell(\bm{X}_i)
+\langle\nabla J_\ell(\bm{X}_i),\bm{P}-\bm{X}_i\rangle_F$.
We choose the permutation minimizing this approximation as a candidate destination $\bm{Q}$. Since $J_\ell(\bm{X}_i)-\langle\nabla J_\ell(\bm{X}_i),\bm{X}_i\rangle_F$ is independent of $\bm{P}$, this reduces to
\begin{align}
\bm{Q}\in\operatorname*{arg\,min}_{\bm{P}\in\mathcal{P}_{d_\ell}}
\langle\nabla J_\ell(\bm{X}_i),\bm{P}\rangle_F.
\label{eq:ro-fw-assignment}
\end{align}
This is a linear assignment problem whose costs are the gradient entries. Since $\bm{Q}$ is a candidate destination, the update direction from $\bm{X}_i$ is $\bm{Q}-\bm{X}_i$.

We then choose $\alpha\in[0,1]$ to minimize the original objective $J_\ell(\bm{X}_i+\alpha(\bm{Q}-\bm{X}_i))$ along this direction. Since this objective is quadratic in $\alpha$, the optimal step size can be computed analytically. The updated iterate, a convex combination of $\bm{X}_i$ and $\bm{Q}$, remains doubly stochastic. We finally discretize the solution and compare it with the permutations encountered during the search.

\Cref{alg:ro-frank-wolfe} in \cref{app:ro-permutation-search} summarizes the procedure.

We next describe how to compute the gradients, objective values, and step sizes used in \cref{alg:ro-frank-wolfe}. The following proposition gives expressions for the gradient and objective value.

\begin{table*}[t]
\centering
\caption{Test accuracy and calibrated test cross-entropy of original and merged models (mean $\pm$ sample standard deviation over three model pairs). Merged models are evaluated at $\lambda=1/2$. Naive merge uses no permutation. The ``Original models'' rows report the average of the two models' metrics. For CNNs, $\times k$ denotes a $k$-fold increase in channel width relative to the standard architecture. Calibration and evaluation details are in \cref{app:merge_evaluation}.}
\label{tab:merge-results}
\small
\setlength{\tabcolsep}{4pt}
\renewcommand{\arraystretch}{1.12}
\begin{tabular}{@{}lll*{6}{c}@{}}
\toprule
\multicolumn{1}{@{}c}{\multirow{3}{*}[-2pt]{Metric}} & \multicolumn{2}{c}{\multirow{3}{*}[-2pt]{Method}} & MNIST & FMNIST & \multicolumn{2}{c}{CIFAR-10} & \multicolumn{2}{c}{CIFAR-100} \\
\cmidrule[0.2pt](lr){4-5}\cmidrule[0.2pt](lr){6-7}\cmidrule[0.2pt](lr){8-9}
 & & & \multicolumn{2}{c}{\multirow{2}{*}{MLP}} & VGG-11 & ResNet-20 & VGG-11 & ResNet-20 \\
 & & & \multicolumn{2}{c}{} & ($\times 1$) & ($\times 16$) & ($\times 2$) & ($\times 32$) \\
\midrule
\multirow{5}{*}[-2pt]{\shortstack[l]{Test\\accuracy\\(\%)}} & \multicolumn{2}{l}{Original models} & $97.58 \pm 0.12$ & $87.87 \pm 0.07$ & $86.72 \pm 0.09$ & $91.88 \pm 0.15$ & $65.59 \pm 0.12$ & $74.49 \pm 0.31$ \\
 & \multicolumn{2}{l}{Naive merge} & $93.42 \pm 1.09$ & $72.51 \pm 9.97$ & $60.92 \pm 6.24$ & $82.84 \pm 1.51$ & $37.88 \pm 6.27$ & $61.43 \pm 2.68$ \\
 & \multicolumn{2}{l}{WM} & $97.58 \pm 0.09$ & $82.90 \pm 4.66$ & $80.42 \pm 3.32$ & $85.32 \pm 1.74$ & $54.18 \pm 1.76$ & $61.38 \pm 2.95$ \\
\cmidrule(l{\tabcolsep}){2-9}
 & \multirow{2}{*}{RO} & Weights & $96.66 \pm 0.23$ & $85.45 \pm 0.69$ & $87.12 \pm 0.65$ & $89.54 \pm 1.00$ & $65.04 \pm 0.23$ & $70.18 \pm 0.45$ \\
 &  & Features & $97.09 \pm 0.62$ & $85.11 \pm 1.54$ & $86.98 \pm 0.40$ & $92.00 \pm 0.44$ & $64.56 \pm 0.35$ & $73.21 \pm 0.74$ \\
\midrule
\multirow{5}{*}[-2pt]{\shortstack[l]{Calibrated\\test loss}} & \multicolumn{2}{l}{Original models} & $0.078 \pm 0.001$ & $0.341 \pm 0.002$ & $0.393 \pm 0.005$ & $0.249 \pm 0.001$ & $1.269 \pm 0.012$ & $0.915 \pm 0.006$ \\
 & \multicolumn{2}{l}{Naive merge} & $0.211 \pm 0.026$ & $0.748 \pm 0.271$ & $1.186 \pm 0.167$ & $0.536 \pm 0.053$ & $2.540 \pm 0.294$ & $1.454 \pm 0.118$ \\
 & \multicolumn{2}{l}{WM} & $0.077 \pm 0.001$ & $0.458 \pm 0.116$ & $0.585 \pm 0.096$ & $0.457 \pm 0.049$ & $1.731 \pm 0.066$ & $1.459 \pm 0.153$ \\
\cmidrule(l{\tabcolsep}){2-9}
 & \multirow{2}{*}{RO} & Weights & $0.107 \pm 0.007$ & $0.403 \pm 0.014$ & $0.374 \pm 0.010$ & $0.325 \pm 0.029$ & $1.291 \pm 0.005$ & $1.096 \pm 0.014$ \\
 &  & Features & $0.093 \pm 0.017$ & $0.410 \pm 0.038$ & $0.382 \pm 0.011$ & $0.249 \pm 0.013$ & $1.310 \pm 0.022$ & $0.974 \pm 0.025$ \\
\bottomrule
\end{tabular}
\end{table*}

\begin{figure*}[t]
\centering
\includegraphics[width=\textwidth]{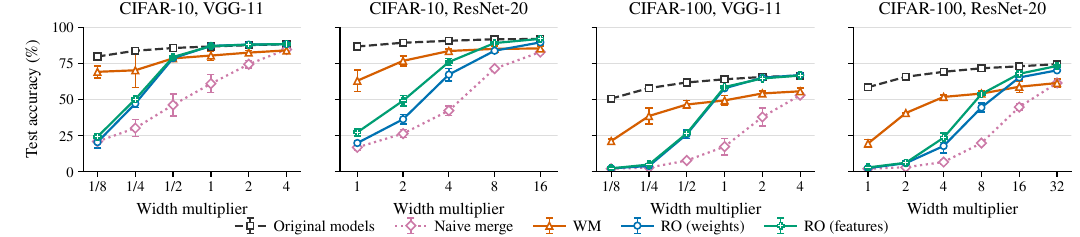}
\caption{Effect of network width on the test accuracy of original and merged models. Merged models use equal-weight parameter averaging ($\lambda=1/2$). Results are shown for VGG-11 and ResNet-20 on CIFAR-10 and CIFAR-100. Error bars indicate sample standard deviations over three model pairs.}
\label{fig:width-comparison}
\label{fig:width-test-accuracy}
\end{figure*}

\begin{proposition}[Gradient and objective value]
\label{prop:ro-gradient-objective}
For the objective $J_\ell$ in \cref{eq:ro-relaxed-objective} and any $\bm{X}\in\mathbb{R}^{d_\ell\times d_\ell}$, we have
\begin{align}
\nabla J_\ell(\bm{X})
&=2\bm{S}_\ell^{(a)}\bm{X}\bm{G}_{\mathrm{col},\ell+1}^{(b)}
+2\bm{G}_{\mathrm{col},\ell+1}^{(a)}\bm{X}\bm{S}_\ell^{(b)}
\label{eq:ro-gradient}\\
J_\ell(\bm{X})&=\tfrac12\langle\bm{X},\nabla J_\ell(\bm{X})\rangle_F.
\label{eq:ro-objective-gradient}
\end{align}

\end{proposition}
The proof is given in \cref{app:proof-ro-gradient-objective}.

We now turn to the step size, which determines how far to move toward $\bm{Q}$ in the update on \cref{line:ro-update} of \cref{alg:ro-frank-wolfe}. Let $\bm{D}=\bm{Q}-\bm{X}$. Along this direction, the objective is
\begin{align}
J_\ell(\bm{X}+\alpha\bm{D})
&=J_\ell(\bm{X})-\alpha g+\alpha^2c,
\label{eq:ro-line-quadratic}
\end{align}
where $g=\langle\nabla J_\ell(\bm{X}),\bm{X}-\bm{Q}\rangle_F\geq0$ is the Frank--Wolfe gap and $c=\tfrac12\langle\bm{D},\nabla J_\ell(\bm{Q})-\nabla J_\ell(\bm{X})\rangle_F=J_\ell(\bm{D})\geq0$ is the quadratic coefficient. The optimal step size over $\alpha\in[0,1]$ is therefore
\begin{align}
\alpha^\star=
\begin{cases}
\min\{1,\max\{0,g/(2c)\}\}, & c>0,\\
0, & c=0
\end{cases}.
\label{eq:ro-fw-step}
\end{align}
If $c=0$ exactly, then $g=0$ as well, and the objective is constant along the linear path. In the implementation, when the curvature is numerically zero, we compare the objective at both endpoints and choose $\alpha=0$ in a tie. After an update, the objective can be computed using \cref{eq:ro-line-quadratic}, and the gradient can be updated by linearity as $(1-\alpha)\nabla J_\ell(\bm{X})+\alpha\nabla J_\ell(\bm{Q})$.

\textbf{Computational complexity.}
As in FAQ \citep{Vogelstein_PLOSONE_2015}, each Frank--Wolfe iteration takes $O(d_\ell^3)$ time.
For layer $\ell$, $k_\ell$ iterations and the final projection require $O((k_\ell+1)d_\ell^3)$ time and $O(d_\ell^2)$ working memory.

\begin{figure*}[t]
\centering
\includegraphics[width=\textwidth]{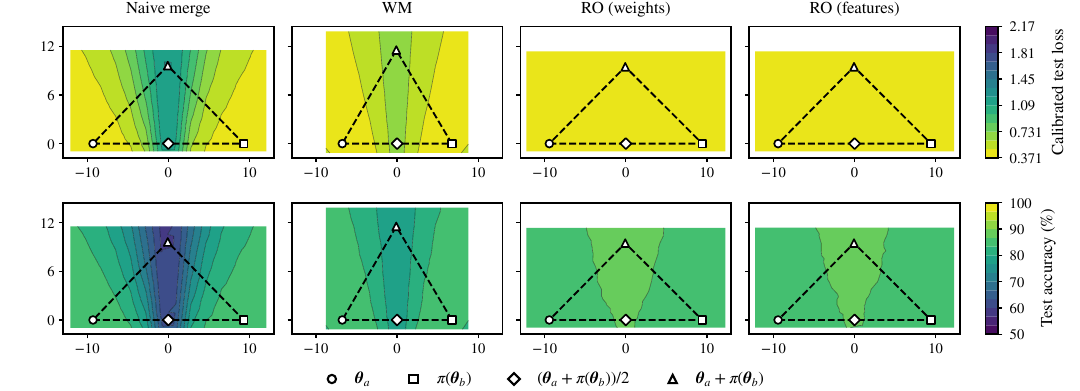}
\caption{Test loss after softmax temperature calibration (top) and test accuracy (bottom) for a single pair of VGG-11 models on CIFAR-10 in a two-dimensional parameter-space plane. Columns compare Naive merge, WM, RO Weights, and RO Features. Markers indicate the two original models, their equal-weight average, and their sum. Dashed lines connect the original models and the sum.}
\label{fig:landscape-cifar10-vgg}
\end{figure*}

\begin{table*}[t]
\centering
\caption[Merging models trained on different datasets]{Merging models trained independently on different datasets, evaluated on the combined test sets.
Values are reported as the mean $\pm$ sample standard deviation over three model pairs.
Naive merge, WM, MuDSC-Align, and ZipIt! average parameters, whereas both RO variants add them. The ensemble baseline averages the original models' logits.
Test loss is uncalibrated cross-entropy. Logit cosine is the mean per-image cosine similarity between merged and ensemble logits.
Dataset splits and label mappings are detailed in \cref{app:cross_dataset_merging}.}
\label{tab:cross-task-merge}
\fontsize{8}{10}\selectfont
\setlength{\tabcolsep}{1.4pt}
\renewcommand{\arraystretch}{1.18}
\begin{tabular*}{\textwidth}{@{\extracolsep{\fill}}ll*{7}{c}@{}}
\toprule
\multirow{2}{*}{Dataset/Model} & \multirow{2}{*}{Metric} & \multirow{2}{*}{Ensemble} & \multirow{2}{*}{Naive merge} & \multirow{2}{*}{WM} & \multirow{2}{*}{MuDSC-Align} & \multirow{2}{*}{ZipIt!} & RO & RO \\
 & & & & & & & (weights) & (features) \\
\midrule
\multirow{3}{*}[-2pt]{\shortstack[l]{MNIST--FMNIST\\MLP}} & Test acc. (\%) & $73.43 \pm 0.85$ & $58.44 \pm 3.30$ & $44.54 \pm 3.98$ & $45.87 \pm 2.38$ & $61.98 \pm 5.30$ & $72.25 \pm 1.24$ & $72.25 \pm 1.31$ \\
 & Test loss & $0.900 \pm 0.051$ & $1.863 \pm 0.024$ & $1.540 \pm 0.075$ & $1.543 \pm 0.135$ & $1.520 \pm 0.086$ & $1.215 \pm 0.155$ & $1.193 \pm 0.150$ \\
 & Logit cosine & --- & $0.893 \pm 0.007$ & $0.791 \pm 0.023$ & $0.822 \pm 0.016$ & $0.929 \pm 0.018$ & $0.994 \pm 0.002$ & $0.996 \pm 0.001$ \\
\midrule
\multirow{3}{*}[-2pt]{\shortstack[l]{Split CIFAR-10\\VGG-11 ($\times1$)}} & Test acc. (\%) & $78.58 \pm 0.13$ & $51.29 \pm 5.04$ & $42.40 \pm 5.99$ & $51.95 \pm 3.16$ & $50.90 \pm 8.05$ & $76.81 \pm 0.80$ & $77.18 \pm 1.12$ \\
 & Test loss & $0.610 \pm 0.007$ & $1.246 \pm 0.097$ & $1.464 \pm 0.037$ & $1.136 \pm 0.062$ & $1.183 \pm 0.167$ & $0.857 \pm 0.052$ & $0.798 \pm 0.063$ \\
 & Logit cosine & --- & $0.588 \pm 0.056$ & $0.495 \pm 0.112$ & $0.693 \pm 0.055$ & $0.641 \pm 0.164$ & $0.902 \pm 0.010$ & $0.915 \pm 0.025$ \\
\midrule
\multirow{3}{*}[-2pt]{\shortstack[l]{Split CIFAR-10\\ResNet-20 ($\times16$)}} & Test acc. (\%) & $79.51 \pm 0.54$ & $60.56 \pm 1.64$ & $55.30 \pm 3.95$ & $65.31 \pm 1.27$ & $60.05 \pm 3.84$ & $78.55 \pm 1.43$ & $83.25 \pm 1.11$ \\
 & Test loss & $0.557 \pm 0.007$ & $1.161 \pm 0.030$ & $1.419 \pm 0.206$ & $0.913 \pm 0.037$ & $1.046 \pm 0.013$ & $0.691 \pm 0.033$ & $0.524 \pm 0.024$ \\
 & Logit cosine & --- & $0.726 \pm 0.032$ & $0.581 \pm 0.035$ & $0.758 \pm 0.041$ & $0.769 \pm 0.012$ & $0.889 \pm 0.014$ & $0.950 \pm 0.005$ \\
\bottomrule
\end{tabular*}
\end{table*}

\section{Experiments}
\label{sec:experiments}

We empirically examine how the permutations found by our RO-based methods affect model connectivity and merging.
We first evaluate whether these permutations promote LEWC between models trained independently on the same dataset (\cref{sec:exp-same-conditions}). We then examine the geometry of the resulting low-loss region (\cref{sec:exp-loss-landscape}) and evaluate model merging across different training datasets (\cref{sec:exp-different-datasets}). In the first two experiments, we evaluate loss after softmax temperature calibration.

\subsection{LEWC through Permutation Search}
\label{sec:exp-same-conditions}


We study LEWC-based connectivity between two models trained independently on the same dataset, using the same architecture and training hyperparameters but different random seeds. We compare parameter averaging without permutation (Naive merge), WM, and the two proposed RO permutation search methods. Our choices of datasets, architecture families, and main training hyperparameters largely follow those of \citet{Ito_ICLR_2026}. Details are given in \cref{app:experimental_setup}.

\Cref{tab:merge-results} reports test accuracy and calibrated test loss for models obtained by parameter interpolation (hereafter, merged models) at $\lambda=1/2$. We choose this coefficient because prior work observed substantial performance degradation near the interpolation midpoint \citep{Ainsworth_ICLR_2023,Ito_ICLR_2026}. The MLPs have 512 units per hidden layer. The width multipliers for VGG-11 and ResNet-20 are $\times1$ and $\times16$ on CIFAR-10, and $\times2$ and $\times32$ on CIFAR-100, respectively. RO Weights uses only weights, whereas RO Features uses features.

Both proposed methods achieve higher accuracy and lower loss than naive merge in every setting, narrowing the performance gap to the original models. The two proposed RO permutation search methods perform similarly on MLPs and VGG-11, whereas RO Features achieves higher accuracy and lower loss on ResNet-20. This suggests that effective permutations can often be found using weights alone. WM preserves the original models' performance for MLPs on MNIST but degrades performance in the other settings, where the proposed methods outperform it.

To examine the effect of network width, we vary the widths of VGG-11 and ResNet-20 and report the resulting test accuracies in \cref{fig:width-comparison}. The corresponding calibrated test losses are shown in \cref{fig:width-calibrated-test-loss}. The models and methods are the same as in \cref{tab:merge-results}. Overall, increasing width improves the merged models' test accuracy and reduces their calibrated test loss. As shown by \citet{Ito_ICLR_2026}, widening also narrows the gap to the original models without permutations, but matching their performance requires substantial width expansion. The proposed method narrows this gap at smaller widths, bringing performance closer to that of the original models. These results support the effectiveness of RO-based permutation search in facilitating LEWC-based LMC with less width expansion.

\subsection{Geometry of the Low-Loss Region}
\label{sec:exp-loss-landscape}


We examine calibrated loss and test accuracy after applying the discovered permutations to characterize the low-loss region beyond the linear path between the original models. \Cref{fig:landscape-cifar10-vgg} shows results for VGG-11 on CIFAR-10. Results for the other five dataset--architecture combinations are provided in \cref{app:loss_landscape}. At each evaluated point, we apply softmax temperature calibration before computing test loss.

Each plot shows performance on the plane containing the triangle with vertices $\bm{\theta}^{(a)}$, $\pi(\bm{\theta}^{(b)})$, and their sum $\bm{\theta}^{(a)}+\pi(\bm{\theta}^{(b)})$, including the region around the triangle. The base corresponds to linear interpolation between the two original models, with its midpoint representing parameter averaging at $\lambda=1/2$. Including the parameter sum allows us to examine whether the low-loss region extends to the other edges and the interior of this triangle.

To explain the role of the parameter sum, we omit biases as in \cref{sec:ro-lmc} and assume that weak additivity and RO hold between the permuted models. Consider the first layer. Under parameter averaging, its activation is $\bm{a}_1^{(c)}=(\bm{a}_1^{(a)}+\widetilde{\bm{a}}_1^{(b)})/2$. The second-layer weights are also averaged, giving $\bm{W}_2^{(c)}=(\bm{W}_2^{(a)}+\widetilde{\bm{W}}_2^{(b)})/2$. Expanding the product and using RO to eliminate the cross terms yields
\begin{align}
\bm{z}_2^{(c)}
&=\frac{\bm{W}_2^{(a)}+\widetilde{\bm{W}}_2^{(b)}}{2}
\frac{\bm{a}_1^{(a)}+\widetilde{\bm{a}}_1^{(b)}}{2}
\\
&=\tfrac14\bigl(\bm{W}_2^{(a)}\bm{a}_1^{(a)}
+\widetilde{\bm{W}}_2^{(b)}\widetilde{\bm{a}}_1^{(b)}\bigr).
\end{align}
The coefficient is $1/4$ because both the features and weights are scaled by $1/2$. Repeating this computation across layers scales the sum of the original activations by $2^{-\ell}$, as in \cref{def:lewc}. With parameter addition, each model's weights have coefficient 1, so the activations are not scaled down and $\bm{a}_\ell^{(c)}=\bm{a}_\ell^{(a)}+\widetilde{\bm{a}}_\ell^{(b)}$ at every layer. The output logits are therefore the sum of the original models' logits, motivating the inclusion of this point in the geometric analysis.

In \cref{fig:landscape-cifar10-vgg}, naive merge exhibits decreased accuracy between the two original models. Consistent with previous findings~\citep{Entezari_arxiv_2022,Ainsworth_ICLR_2023}, WM improves accuracy along this direct linear interpolation. The permutations promoting RO yield low calibrated loss and high accuracy throughout the triangle, including its interior. The other dataset--architecture combinations in \cref{app:loss_landscape} show similar trends.

These results extend the geometric picture of LMC from a low-loss linear path to a low-loss region containing both original models and their parameter sum. The inclusion of the sum, together with its logit-sum interpretation under the conditions for LEWC, motivates using parameter addition for model merging in the next experiment.

\subsection{Merging Models Trained on Different Datasets}
\label{sec:exp-different-datasets}


We evaluate merging models trained on different data distributions. Specifically, we consider MLPs trained separately on MNIST and Fashion-MNIST, and VGG-11 and ResNet-20 models trained on disjoint CIFAR-10 subsets containing classes 0--4 and 5--9. Each CIFAR-10 model is a five-class classifier, with classes 5--9 remapped to labels 0--4. The merged model thus retains five outputs. For example, the first output corresponds to both original classes 0 and 5. The MNIST and Fashion-MNIST models retain their original numerical labels and ten outputs. Details are given in \cref{app:cross_dataset_merging}.

In addition to naive merge and WM, we compare against MuDSC-Align \citep{Xu_CVPR_2024} and ZipIt! \citep{Stoica_ICLR_2024}. MuDSC-Align matches neurons using a combination of weight and activation similarities, whereas ZipIt! merges features both within and across models. Implementation details and hyperparameters are given in \cref{app:cross_dataset_merging}.

\Cref{tab:cross-task-merge} reports the results. The ensemble baseline averages the logits of the two original models. Naive merge, WM, MuDSC-Align, and ZipIt! use parameter averaging. Motivated by the logit-sum property and low-loss geometry discussed in \cref{sec:exp-loss-landscape}, our methods add the parameters after permutation. Test accuracy and test loss are measured on the concatenation of the two corresponding test sets, with loss computed as cross-entropy without temperature calibration. Logit cosine is the cosine similarity between the merged model's logits and the ensemble's logits, computed per input and then averaged.

The proposed methods achieve test accuracy close to that of the ensemble in all settings. In particular, RO Features achieves 83.25\% accuracy on ResNet-20, exceeding the ensemble's 79.51\%. In the other cases, accuracy is slightly below that of the ensemble but substantially higher than that of any merging baseline. The proposed methods also achieve lower test loss than all merging baselines in every setting, with values comparable to those of the ensemble.

The proposed methods achieve higher logit cosine similarity than all merging baselines in all settings, exceeding 0.99 for both RO variants on MLPs. This indicates that our methods approximately reproduce the direction of the logit ensemble's outputs even when merging models trained on different distributions. In contrast, WM has lower accuracy than naive merge in every setting, illustrating its difficulty in merging models trained on different distributions under these conditions. These results suggest that suppressing interference through RO can integrate models with different functions into a single network.

A practical benefit of this integration is that ensemble-like outputs can be obtained with a single forward pass, even when the original models were trained on different datasets. Further fine-tuning could improve performance at a lower cost than training a new model from scratch. Another possible extension is to use separate output heads with shared merged hidden layers to help preserve each original model's performance on its own task. We leave these extensions to future work.

\section{Conclusion}
\label{sec:conclusion}

We proposed weight-only and feature-based permutation searches that promote LEWC by encouraging RO. Experiments showed that both methods facilitate LMC under temperature calibration with less width expansion than permutation-free interpolation, including VGG-11 on CIFAR-10 without width expansion. Loss-landscape analysis revealed low calibrated loss throughout the triangle spanned by the two models and their parameter sum after permutation. This extends the geometric picture of connectivity beyond the linear path between the original models. The methods also successfully merged models trained on different data distributions, yielding predictions close to those of a logit ensemble.

Future work includes further reducing width requirements and extending the approach beyond image classification to domains such as natural language processing. Recent work generalizing weight matching and LMC to Transformers \citep{Theus_NIPS_2025} also motivates adapting our approach to these architectures.

\subsection*{AI Use Statement}
We used generative AI tools to assist with English writing and editing, code development, and implementation-related discussions, including assistance with implementing RO-based permutation search. The authors reviewed all AI-assisted text, code, and technical suggestions and made the final decisions regarding their use. The authors take full responsibility for the content of the paper, the implementation, and the reported results.

\bibliography{modified}

\section*{Checklist}

\begin{enumerate}

  \item For all models and algorithms presented, check if you include:
  \begin{enumerate}
    \item A clear description of the mathematical setting, assumptions, algorithm, and/or model. [Yes] See \cref{sec:preliminaries,sec:proposed_method}.
    \item An analysis of the properties and complexity (time, space, sample size) of any algorithm. [Yes] See \cref{sec:ro-frank-wolfe}.
    \item (Optional) Anonymized source code, with specification of all dependencies, including external libraries. [Yes] See \cref{app:experimental_setup,app:computational_environment}.
  \end{enumerate}

  \item For any theoretical claim, check if you include:
  \begin{enumerate}
    \item Statements of the full set of assumptions of all theoretical results. [Yes] See \cref{sec:ro-lmc,sec:proposed_method,app:cnn_resnet}.
    \item Complete proofs of all theoretical results. [Yes] See \cref{app:proofs,app:cnn_resnet}.
    \item Clear explanations of any assumptions. [Yes] See \cref{sec:ro-lmc,sec:proposed_method,app:cnn_resnet}.
  \end{enumerate}

  \item For all figures and tables that present empirical results, check if you include:
  \begin{enumerate}
    \item The code, data, and instructions needed to reproduce the main experimental results (either in the supplemental material or as a URL). [Yes] See \cref{app:experimental_setup}.
    \item All the training details (e.g., data splits, hyperparameters, how they were chosen). [Yes] See \cref{app:model_training,app:permutation_search,app:cross_dataset_merging}.
    \item A clear definition of the specific measure or statistics and error bars (e.g., with respect to the random seed after running experiments multiple times). [Yes] See \cref{app:merge_evaluation,app:cross_dataset_merging} and the figure and table captions.
    \item A description of the computing infrastructure used. (e.g., type of GPUs, internal cluster, or cloud provider). [Yes] See \cref{app:computational_environment}.
  \end{enumerate}

  \item If you are using existing assets (e.g., code, data, models) or curating/releasing new assets, check if you include:
  \begin{enumerate}
    \item Citations of the creator If your work uses existing assets. [Yes] See \cref{app:experimental_setup}.
    \item The license information of the assets, if applicable. [Yes] See \cref{app:model_training,app:computational_environment,app:cross_dataset_merging}.
    \item New assets either in the supplemental material or as a URL, if applicable. [Yes] See the anonymous code repository linked in \cref{app:experimental_setup}.
    \item Information about consent from data providers/curators. [Not Applicable]
    \item Discussion of sensible content if applicable, e.g., personally identifiable information or offensive content. [Not Applicable]
  \end{enumerate}

  \item If you used crowdsourcing or conducted research with human subjects, check if you include:
  \begin{enumerate}
    \item The full text of instructions given to participants and screenshots. [Not Applicable]
    \item Descriptions of potential participant risks, with links to Institutional Review Board (IRB) approvals if applicable. [Not Applicable]
    \item The estimated hourly wage paid to participants and the total amount spent on participant compensation. [Not Applicable]
  \end{enumerate}

\end{enumerate}

\clearpage
\appendix
\thispagestyle{empty}
\onecolumn
\aistatstitle{Supplementary Materials: Permutation-Based Pre-trained Model Merging via Reciprocal Orthogonality}
\section{Related Work}
\label{app:related_work}

\subsection{(Linear) Mode Connectivity}
\label{app:related_mode_connectivity}

Mode connectivity refers to the existence of a continuous low-loss path in parameter space connecting two trained models.
\citet{Garipov_NIPS_2018} and \citet{Draxler_ICML_2018} demonstrated mode connectivity between independently trained networks by constructing simple low-loss curves.
\citet{Benton_ICML_2021} extended the study of mode connectivity from paths to multidimensional low-loss regions and used these regions for ensembling.
Linear mode connectivity (LMC) holds when loss remains low along the linear path connecting the two models' parameters.
\citet{frankle_ICML_2020} studied LMC between models trained from a shared checkpoint with different data orders and augmentations.
They used error barriers along the linear interpolation to measure stability to SGD randomness.
They found that sparse subnetworks obtained by iterative magnitude pruning reached the full network's accuracy only when LMC held between their trained copies, linking this stability to the lottery ticket hypothesis.

For networks trained from different initializations, permutation symmetries can help explain LMC.
\citet{Entezari_arxiv_2022} conjectured that independently trained solutions lie in a single approximately convex low-loss basin after appropriate permutations.
\citet{Ainsworth_ICLR_2023} provided empirical support for this conjecture by demonstrating LMC after permutation search.
Their methods include WM, which finds permutations that reduce the distance between the models' weights.

Subsequent work has extended the study of LMC to multiple models, asking whether permutations can place them in a single approximately convex low-loss region.
\citet{Crisostom_arXiv_2024} studied LMC among multiple models by extending the weight-matching approach to search for permutations that minimize pairwise weight distances while satisfying cycle consistency.
Their method uses Frank--Wolfe to optimize permutations jointly across layers.
\citet{Ito_ICML_2025} proposed STE-MM, a permutation search method based on a straight-through estimator, to promote LMC among multiple models.
They showed empirically that the aligned models have small pairwise loss barriers and that their parameter average has low loss.

Other studies have investigated how LMC is related to the representations learned by each model.
\citet{Zhou_NIPS_2023} introduced layerwise linear feature connectivity to explain LMC through the relationship between the hidden representations of the original models and those of the interpolated model.
\citet{Ito_ICLR_2025} analyzed how WM promotes LMC by aligning dominant singular vectors of weight matrices and increasing similarity between the models' hidden-layer functions.
\citet{Lubana_ICML_2023} studied the relationship between mode connectivity and similarity in the models' predictive mechanisms, using invariance to input transformations to characterize these mechanisms.
In this work, we study LMC through LEWC, where the two models' computations proceed approximately independently within the merged network.

\subsection{Model Merging}
\label{app:related_model_merging}

Many methods for merging independently trained networks first find matching neurons and then combine the corresponding parameters.
\citet{Singh_NIPS_2020} used optimal transport to align neurons for model fusion.
\citet{Liu_ICML_2022} proposed a model merging method based on graph matching.
Their method formulates neuron matching as a QAP based on weight similarity and network structure.
\citet{Xu_CVPR_2024} showed that neurons with similar weights can have different activations.
Their MuDSC method uses a weighted combination of weight and activation similarities to match neurons and merge pretrained models without additional training.
To retain useful features when merging models trained on different tasks, ZipIt \citep{Stoica_ICLR_2024} merges features both within and across models and can leave later layers separate.
PLeaS \citep{Nasery_CVPR_2025} uses partial feature alignment to control the size of the merged network and estimates its weights by least squares.
In contrast to these similarity-based approaches, our method searches for permutations that promote RO between one model's weights and the other model's features.
We formulate this search as a QAP and show that reducing interference in this way enables model merging.

Another line of work merges models fine-tuned from the same pretrained model.
Model soups \citep{Wortsman_ICML_2022} average the weights of models fine-tuned with different hyperparameters.
Fisher merging \citep{Matena_NeurIPS_2022} uses the Fisher information to weight the contribution of each parameter.
TIES-Merging \citep{Yadav_NeurIPS_2023} removes small parameter updates and resolves disagreements in their signs to reduce interference during merging.
\citet{Gargiulo_CVPR_2025} used the singular vectors of parameter update matrices to reduce interference between tasks.
In this work, we consider models trained from different initializations and use permutations to reduce interference between one model's weights and the other's features.

Theoretical studies also ask when merged models can preserve the predictions of the original models.
\citet{Kuzborskij_AISTATS_2026} showed that adding the weights of separately trained models can approximately preserve their predictions on their respective datasets.
Their analysis covers shallow ReLU and deep linear networks under assumptions that include approximate orthogonality between inputs from different datasets and weight decay during training.
The orthogonality we target is between learned weights and hidden features, and we promote it through permutation search.

\subsection{Effect of Model Width on Loss Landscape}
\label{app:related_model_width}

Increasing network width can change the geometry of the loss landscape and improve connectivity between solutions.
\citet{Nguyen_ICML_2019} proved that loss sublevel sets are connected when the network is sufficiently wide, under assumptions on the architecture and activation functions.
\citet{Simsek_ICML_2021} showed that adding neurons can connect zero-loss minima related by neuron permutations.
\citet{Kim_ICLR_2025} studied how width affects the structure and connectivity of the set of optimal solutions in regularized two-layer networks.
These results show that widening can improve connectivity between solutions.
However, they do not generally guarantee low loss along the linear path between two trained models.

Width also affects the connectivity of solutions found by training algorithms.
\citet{Shevchenko_ICML_2020} used dropout stability to show that, under their assumptions, SGD solutions can be connected by piecewise linear paths with loss barriers that decrease as width increases.
\citet{Nguyen_NeurIPS_2021} used properties of intermediate features to establish connectivity under weaker width requirements.
\citet{Zhan_AISTATS_2025} analyzed how width affects loss barriers after permutation in a two-layer ReLU teacher--student setting.

\citet{Ito_ICLR_2026} showed empirically that sufficiently wide networks can exhibit LMC after softmax temperature calibration without permutations.
Their LEWC framework describes how the merged model produces outputs proportional to a weighted logit ensemble of the original models.
RO and ReLU weak additivity are jointly sufficient for LEWC.
Motivated by this result, we investigate whether permutations that promote RO can lead to LEWC and calibrated LMC at smaller widths.

\section{Experimental Setup}
\label{app:experimental_setup}

This section provides details of the experimental setup.
We describe the datasets and model training in \cref{app:model_training}, permutation search in \cref{app:permutation_search}, and merged-model evaluation in \cref{app:merge_evaluation}.
We then describe the computational environment in \cref{app:computational_environment} and the setup for merging models trained on different datasets in \cref{app:cross_dataset_merging}.

Our implementation is available in the following repository.\footnote{\texttt{[URL]}}

\subsection{Datasets and Model Training}
\label{app:model_training}

We use the same benchmarks, architecture families, optimizers, learning rates, weight decay, and batch sizes as \citet[Appendix~B]{Ito_ICLR_2026}.

\paragraph{Datasets and model pairs.}
We evaluate MLPs on MNIST \citep{MNIST} and Fashion-MNIST (FMNIST) \citep{FMNIST}\footnote{Fashion-MNIST is distributed under the \href{https://github.com/zalandoresearch/fashion-mnist/blob/master/LICENSE}{MIT License}.}, and VGG-11 and ResNet-20 on CIFAR-10 and CIFAR-100 \citep{CIFAR10}.
For each dataset, architecture, and width configuration, we train six models with different random seeds and group them into three pairs.
All methods use the same model pairs.

\paragraph{Data splits and checkpoint selection.}
We reserve approximately $10\%$ of the training examples in each class for validation.
The resulting training/validation sizes are 54,004/5,996 for MNIST, 54,000/6,000 for FMNIST, and 45,000/5,000 for each CIFAR dataset.
The MNIST split sizes are not round numbers because its class sizes are unequal.
We train MLPs for 50 epochs and CNNs for 100 epochs to minimize cross-entropy, without early stopping.
We retain the checkpoint with the lowest validation loss.

\paragraph{Architectures.}
Each MLP receives a flattened $28\times28$ image and has three hidden layers of 512 units with ReLU activations.
VGG-11 \citep{Simonyan_ICLR_2015} is adapted to $32\times32$ inputs, with BatchNorm after each convolution and a single linear classifier.
ResNet-20 \citep{He_CVPR_2016} has three stages of three residual blocks, with base channel widths of 16, 32, and 64.
Stage transitions use $3\times3$ convolutional projection shortcuts with stride 2.

\paragraph{Optimization.}
\Cref{tab:training-hyperparameters} summarizes the training hyperparameters.
Adam uses $(\beta_1,\beta_2)=(0.9,0.999)$ and $\epsilon=10^{-8}$, while SGD uses momentum 0.9.
For CNNs, the learning rate increases linearly from zero to 0.4 over five warm-up epochs and then decays linearly toward zero over the remaining epochs.

\begin{table}[bp]
\centering
\caption{Training hyperparameters. The CNN settings apply to both VGG-11 and ResNet-20 on both CIFAR datasets.}
\label{tab:training-hyperparameters}
\small
\begin{tabular}{@{}lcc@{}}
\toprule
Setting & MLP & CNNs \\
\midrule
Optimizer & Adam & SGD \\
Epochs & 50 & 100 \\
Batch size & 512 & 500 \\
Learning rate & $10^{-3}$ & $0.4$ (peak) \\
Weight decay & $3\times10^{-3}$ & $3\times10^{-3}$ \\
Learning-rate schedule & Constant & Linear warm-up and decay \\
\bottomrule
\end{tabular}
\end{table}

\paragraph{Preprocessing.}
MNIST and FMNIST pixel intensities are scaled to $[0,1]$, without augmentation.
For CIFAR training, we use random horizontal flips and random $32\times32$ crops with four pixels of padding.
For CIFAR-10, the channel means and standard deviations used for normalization are $(0.485,0.456,0.406)$ and $(0.229,0.224,0.225)$, respectively. For CIFAR-100, they are $(0.5071,0.4867,0.4408)$ and $(0.2675,0.2565,0.2761)$.
Validation and test inputs use the same normalization without random augmentation.

\subsection{Permutation Search}
\label{app:permutation_search}

\paragraph{Activation statistics for RO Features.}
We uniformly sample $1\%$ of the training set, excluding validation data, yielding 540 images for each of MNIST and FMNIST, and 450 for each CIFAR dataset.
We obtain activations from the original models in evaluation mode without data augmentation.
We compute the second moments without subtracting the mean and normalize them by the number of images.

\paragraph{Frank--Wolfe optimization.}
We initialize each distinct permutation variable to the identity and optimize it independently using the procedure in \cref{sec:ro-frank-wolfe}, with $\varepsilon=10^{-7}$ and $n_{\mathrm{check}}=500$.
Every $n_{\mathrm{check}}$ updates, we recompute the objective and divide the total improvement over the last $n_{\mathrm{check}}$ updates by $n_{\mathrm{check}}|J_\ell(\bm{I}_{d_\ell})|$. The search terminates if this normalized improvement is at most $\varepsilon$. It also terminates when the objective is sufficiently small or the step size is zero.
Linear assignment problems are solved on the CPU using LAPJV.
To accelerate repeated solves, we reuse dual variables from the preceding solve to adjust the cost matrix, following \citet{Ito_ICML_2025}.

\paragraph{Weight matching.}
We use a PyTorch implementation of the weight matching (WM) method of \citet{Ainsworth_ICLR_2023}.
Starting from identity permutations, WM updates one permutation variable at a time while keeping the others fixed.
Each sweep updates every permutation variable once in a randomized order.
We allow at most 300 sweeps and stop earlier if a full sweep produces no improvement above the numerical tolerance.

\subsection{Evaluation of Merged Models}
\label{app:merge_evaluation}

The main comparison uses equal-weight parameter averaging ($\lambda=1/2$), with and without permutation alignment.
Following \citet{Jordan_ICLR_2023}, we reset and recompute the running statistics of each merged model with BatchNorm.
We use one pass over the training data, with batch size 500 and no data augmentation.

For calibrated losses, we fit a positive scalar inverse temperature separately for each original and merged model by minimizing validation cross-entropy, and apply it to the test logits.
All metrics use the 10,000 test examples for each dataset.
We report the mean and sample standard deviation across the three model pairs, except for the loss-landscape plots, which use a single pair for each dataset--architecture configuration.

\subsection{Computational Environment}
\label{app:computational_environment}

We run experiments on eight NVIDIA RTX 6000 Ada GPUs.
The software environment consists of Python 3.12.9, PyTorch 2.11.0 \citep{Paszke_NeurIPS_2019} with CUDA 12.8, Lightning 2.6.6 \citep{Falcon_Lightning_2019}, torchvision 0.26.0 \citep{TorchVision_2016}, and \texttt{lapjv} 1.3.27 \citep{Markovtsev_LAPJV_2024}.
Training uses bfloat16 mixed precision with deterministic cuDNN operations enabled.
\href{https://github.com/pytorch/pytorch/blob/v2.11.0/LICENSE}{PyTorch} and \href{https://github.com/pytorch/vision/blob/v0.26.0/LICENSE}{torchvision} are distributed under the BSD 3-Clause License, \href{https://github.com/Lightning-AI/pytorch-lightning/blob/2.6.6/LICENSE}{Lightning} under the Apache License 2.0, and \href{https://pypi.org/project/lapjv/1.3.27/}{\texttt{lapjv}} under the MIT License.

\subsection{Merging Models Trained on Different Datasets}
\label{app:cross_dataset_merging}

For \cref{tab:cross-task-merge}, models are trained separately on MNIST and Fashion-MNIST, or on disjoint CIFAR-10 subsets containing classes 0--4 and 5--9.
The MNIST and Fashion-MNIST models both have ten output coordinates and retain their original numerical labels. The same label index therefore represents different classes in the two datasets.
The split CIFAR-10 models each have five output coordinates. Labels 0--4 are retained for the first subset, and classes 5--9 are remapped to labels 0--4 for the second.
Evaluation uses the union of the two test sets with these label mappings, without expanding the output space.
We train six models with different random seeds and form three pairs, each containing one model trained on each dataset.

Naive merge, WM, MuDSC-Align, and ZipIt! use parameter averaging, whereas both RO variants use parameter addition after permutation search.
The ensemble averages the two original models' raw logits.
We evaluate cross-entropy without temperature calibration and average the per-image cosine similarity between merged and ensemble logits.
Cosine similarity measures their directional agreement and is invariant to positive rescaling.

\paragraph{MuDSC-Align and ZipIt!}
We use the authors' implementations of MuDSC-Align \citep{Xu_CVPR_2024}\footnote{\url{https://github.com/zju-vipa/training_free_model_merging}} and ZipIt! \citep{Stoica_ICLR_2024}\footnote{\url{https://github.com/gstoica27/ZipIt}. The ZipIt! implementation is distributed under the \href{https://github.com/gstoica27/ZipIt/blob/master/LICENSE}{MIT License}.}.
For both methods, we extract the required features using all training data.
For MuDSC-Align, we select the coefficient $\alpha$, which balances weight and activation similarities, from $\{0,0.1,\ldots,1\}$ to maximize the merged model's accuracy on the validation set.
For ZipIt!, we use full merging and select $(\alpha,\beta)$ separately for each model pair to maximize the merged model's accuracy on the validation set.
Following the authors' public GitHub implementation, we perform a $6\times6$ grid search over $\alpha\in\{0.0001,0.01,0.1,0.3,0.5,1\}$ and $\beta\in\{0,0.0375,0.075,0.125,0.25,1\}$.

\FloatBarrier
\section{Details of Permutation Search}
\label{app:ro-permutation-search}

This section gives the pseudocode for the permutation search described in \cref{sec:ro-frank-wolfe}.

\begin{algorithm}[tbp]
\caption{RO permutation search for one hidden layer}
\label{alg:ro-frank-wolfe}
\begin{algorithmic}[1]
\Require Objective $J_\ell$, width $d_\ell$, check interval $n_{\mathrm{check}}$, tolerance $\varepsilon$
\Ensure Approximate permutation solution $\widehat{\bm{P}}_\ell\in\mathcal{P}_{d_\ell}$
\State $\bm{X}\gets\bm{I}_{d_\ell}$, $\bm{P}_{\mathrm{best}}\gets\bm{I}_{d_\ell}$\label{line:ro-init}
\While{\textbf{True}}
  \State $\bm{Q}\gets\operatorname*{arg\,min}_{\bm{P}\in\mathcal{P}_{d_\ell}}\langle\nabla J_\ell(\bm{X}),\bm{P}\rangle_F$\label{line:ro-assignment}
  \State $\bm{P}_{\mathrm{best}}\gets\operatorname*{arg\,min}_{\bm{P}\in\{\bm{P}_{\mathrm{best}},\bm{Q}\}}J_\ell(\bm{P})$\label{line:ro-best}
  \State Compute $\alpha$ using \cref{eq:ro-fw-step}\label{line:ro-step}
  \If{$\alpha=0$}\label{line:ro-zero-step}
    \State \textbf{break}
  \EndIf
  \State $\bm{X}\gets(1-\alpha)\bm{X}+\alpha\bm{Q}$\label{line:ro-update}
  \If{the update count is a multiple of $n_{\mathrm{check}}$}\label{line:ro-check-interval}
    \State Recompute $J_\ell(\bm{X})$
    \If{a stopping condition is met}\label{line:ro-stop}
      \State \textbf{break}\label{line:ro-stop-break}
    \EndIf
  \EndIf
\EndWhile
\State $\bm{Q}_{\mathrm{proj}}\gets\operatorname*{arg\,max}_{\bm{P}\in\mathcal{P}_{d_\ell}}\langle\bm{X},\bm{P}\rangle_F$\label{line:ro-projection}
\State \Return $\operatorname*{arg\,min}_{\bm{P}\in\{\bm{P}_{\mathrm{best}},\bm{Q}_{\mathrm{proj}}\}}J_\ell(\bm{P})$\label{line:ro-return}
\end{algorithmic}
\end{algorithm}

\Cref{alg:ro-frank-wolfe} summarizes this procedure. It initializes at the identity (\cref{line:ro-init}), selects and compares permutations (\cref{line:ro-assignment,line:ro-best}), and updates the relaxed iterate (\crefrange{line:ro-step}{line:ro-update}). It periodically checks the stopping conditions (\crefrange{line:ro-check-interval}{line:ro-stop-break}), which are detailed in \cref{app:permutation_search}. Finally, it compares the nearest permutation in Frobenius distance with the best stored candidate (\crefrange{line:ro-projection}{line:ro-return}). Independent searches over hidden layers yield $\{\bm{P}_\ell\}_{\ell=1}^{L-1}$, defining $\pi$.

\FloatBarrier
\section{Proofs}
\label{app:proofs}

\subsection{Proof of \texorpdfstring{\cref{prop:ro-gradient-objective}}{the Gradient and Objective Proposition}}
\label{app:proof-ro-gradient-objective}

\begin{proof}
Let $\mathrm{d}\bm{X}$ be the matrix of infinitesimal changes in the entries of $\bm{X}$, and let $\mathrm{d}J_\ell$ be the corresponding first-order change in the objective (its total differential). We first derive a general trace identity. For symmetric matrices $\bm{A}$ and $\bm{B}$, consider $F(\bm{X})=\operatorname{tr}(\bm{A}\bm{X}\bm{B}\bm{X}^\top)$. The product rule gives
\begin{align}
\mathrm{d}F
=\operatorname{tr}(\bm{A}\,\mathrm{d}\bm{X}\,\bm{B}\bm{X}^\top)
+\operatorname{tr}(\bm{A}\bm{X}\bm{B}\,\mathrm{d}\bm{X}^\top)
=2\langle\bm{A}\bm{X}\bm{B},\mathrm{d}\bm{X}\rangle_F.
\end{align}
The last equality follows because symmetry of $\bm{A}$ and $\bm{B}$, together with invariance of the trace under transposition and cyclic permutation, makes both terms equal to $\langle\bm{A}\bm{X}\bm{B},\mathrm{d}\bm{X}\rangle_F$. By cyclic permutation, the second term of \cref{eq:ro-relaxed-objective} can be written as $\operatorname{tr}(\bm{S}_\ell^{(a)}\bm{X}\bm{G}_{\mathrm{col},\ell+1}^{(b)}\bm{X}^\top)$. Applying the identity to each term of $J_\ell(\bm{X})$ yields
\begin{align}
\mathrm{d}J_\ell
&=2\langle\bm{G}_{\mathrm{col},\ell+1}^{(a)}\bm{X}\bm{S}_\ell^{(b)},\mathrm{d}\bm{X}\rangle_F
+2\langle\bm{S}_\ell^{(a)}\bm{X}\bm{G}_{\mathrm{col},\ell+1}^{(b)},\mathrm{d}\bm{X}\rangle_F.
\end{align}
Comparing this with $\mathrm{d}J_\ell=\langle\nabla J_\ell(\bm{X}),\mathrm{d}\bm{X}\rangle_F$ gives \cref{eq:ro-gradient}.

The objective value can also be computed from this gradient. Each term of $J_\ell$ contains two factors of $\bm{X}$, so $J_\ell(t\bm{X})=t^2J_\ell(\bm{X})$ for any scalar $t$. With $\bm{X}$ fixed, differentiating both sides with respect to $t$ and applying the chain rule gives
\begin{align}
\frac{\mathrm{d}}{\mathrm{d}t}J_\ell(t\bm{X})
&=\langle\nabla J_\ell(t\bm{X}),\bm{X}\rangle_F
=2tJ_\ell(\bm{X}).
\end{align}
Setting $t=1$ shows that the Frobenius inner product of the gradient and $\bm{X}$ equals twice the objective value, establishing \cref{eq:ro-objective-gradient}.
\end{proof}

\FloatBarrier
\section{Application to Convolutional Networks and Residual Connections}
\label{app:cnn_resnet}

This section explains how to apply our method to convolutional layers. We then discuss the relationship between RO and ensemble-like behavior in networks with residual connections.

\subsection{Consecutive Convolutional Layers}
\label{app:cnn_consecutive}

We consider convolutions with standard zero padding and omit biases. We first describe the matrix representation of convolutional layers, then present the objectives for RO Weights and RO Features, and finally explain the computational procedure for permutation search.

\subsubsection{Matrix Representation of Convolutional Layers}
\label{app:cnn_matrix}

We first express a convolutional layer as a matrix--vector product, as for a fully connected layer. Let $\bm{A}\in\mathbb{R}^{d_{\mathrm{in}}\times h_{\mathrm{in}}\times w_{\mathrm{in}}}$ be the input to a convolutional layer, and let $\bm{Z}\in\mathbb{R}^{d_{\mathrm{out}}\times h_{\mathrm{out}}\times w_{\mathrm{out}}}$ be its output before ReLU. Each channel has a two-dimensional feature map, and stacking these maps along the channel dimension gives the three-dimensional tensors $\bm{A}$ and $\bm{Z}$. Here, $d_{\mathrm{in}},d_{\mathrm{out}}$ are the channel counts, and $h_{\mathrm{in}},h_{\mathrm{out}}$ and $w_{\mathrm{in}},w_{\mathrm{out}}$ are the heights and widths, respectively. Let $n_{\mathrm{in}}=h_{\mathrm{in}}w_{\mathrm{in}}$ and $n_{\mathrm{out}}=h_{\mathrm{out}}w_{\mathrm{out}}$ be the numbers of spatial positions. We temporarily omit the layer and model indices to describe a single layer.

We index channels and spatial positions from $0$ in this section. Let $\operatorname{vec}$ be the operation that reads each channel row by row and concatenates the results in channel order. Specifically,
\begin{align}
[\operatorname{vec}(\bm{A})]_{c n_{\mathrm{in}}+i w_{\mathrm{in}}+j}=A_{c,i,j},
\qquad
0\leq c<d_{\mathrm{in}},\quad 0\leq i<h_{\mathrm{in}},\quad 0\leq j<w_{\mathrm{in}}.
\label{eq:conv-vec-definition-draft}
\end{align}
For example, we read row $0$ of channel $0$ from left to right, continue to the next row, and move to channel $1$ after reading all rows. We apply $\operatorname{vec}$ to $\bm{Z}$ in the same order. Below, we construct a matrix $\bm{T}$ that represents convolution under this vectorization as $\operatorname{vec}(\bm{Z})=\bm{T}\operatorname{vec}(\bm{A})$.

We next define the correspondence between kernel positions and spatial positions. For simplicity, consider a square $k\times k$ kernel. Let $s$ and $\delta$ be the stride and dilation, respectively, shared across the height and width dimensions, and let $p$ be the zero-padding width on each side. Both $s$ and $\delta$ are positive integers, and $p$ is a nonnegative integer. We use $\delta$ for dilation because $d$ denotes the channel count.

Assign each kernel position a scalar index $u$ from $\mathcal{K}=\{0,1,\ldots,k^2-1\}$. Let $r_u=\lfloor u/k\rfloor$ and $t_u=u-k r_u$ be the corresponding row and column within the kernel. For example, a $3\times3$ kernel has $\mathcal{K}=\{0,\ldots,8\}$, where $u=4$ denotes the center $(r_u,t_u)=(1,1)$ and $u=5$ denotes the position $(1,2)$ immediately to its right. Let $\bm{W}_u\in\mathbb{R}^{d_{\mathrm{out}}\times d_{\mathrm{in}}}$ be the channel transformation matrix at position $u$. Its entry $(\bm{W}_u)_{o,c}$ is the weight from input channel $c$ to output channel $o$ at kernel position $(r_u,t_u)$.

We represent the spatial positions accessed by this convolution using a matrix. At output position $(i,j)$, kernel position $u$ accesses input position $(s i-p+\delta r_u,\ s j-p+\delta t_u)$. Let $\bm{B}_u\in\{0,1\}^{n_{\mathrm{out}}\times n_{\mathrm{in}}}$ be the $0$--$1$ matrix representing this correspondence, defined by
\begin{align}
(\bm{B}_u)_{i w_{\mathrm{out}}+j,\ i' w_{\mathrm{in}}+j'}
&=\begin{cases}
1, & i'=s i-p+\delta r_u\ \text{and}\ j'=s j-p+\delta t_u,\\
0, & \text{otherwise}.
\end{cases}
\label{eq:conv-spatial-selector-draft}
\end{align}
Here, $0\leq i<h_{\mathrm{out}}$, $0\leq j<w_{\mathrm{out}}$, $0\leq i'<h_{\mathrm{in}}$, and $0\leq j'<w_{\mathrm{in}}$. Each row of $\bm{B}_u$ corresponds to one output position and contains at most one entry equal to $1$. If the accessed position lies within the input image, only the column corresponding to that input position has value $1$. If it lies in the padding region, no $(i',j')$ satisfies the conditions, so the entire row is $0$. The matrix $\bm{B}_u$ contains no channel weights and depends only on the spatial configuration of the convolution.

\begin{proposition}[Matrix representation of convolution]
\label{prop:conv-matrix-draft}
Under the vectorization above, convolution without bias can be expressed using $\bm{W}_u,\bm{B}_u$ as
\begin{align}
\operatorname{vec}(\bm{Z})=\bm{T}\operatorname{vec}(\bm{A}),
\qquad
\bm{T}=\sum_{u\in\mathcal{K}}\bm{W}_u\otimes\bm{B}_u.
\label{eq:conv-kronecker-draft}
\end{align}
Here, $\otimes$ denotes the Kronecker product, and $\bm{T}\in\mathbb{R}^{d_{\mathrm{out}}n_{\mathrm{out}}\times d_{\mathrm{in}}n_{\mathrm{in}}}$.
\end{proposition}
\begin{proof}
Set $A_{c,i',j'}=0$ outside the input image. Expanding one output entry and applying \cref{eq:conv-spatial-selector-draft} gives
\begin{align}
Z_{o,i,j}
&=\sum_{c=0}^{d_{\mathrm{in}}-1}\sum_{u\in\mathcal{K}}(\bm{W}_u)_{o,c}\,A_{c,s i-p+\delta r_u,s j-p+\delta t_u}
\nonumber\\
&=\sum_{c=0}^{d_{\mathrm{in}}-1}\sum_{u\in\mathcal{K}}\sum_{i'=0}^{h_{\mathrm{in}}-1}\sum_{j'=0}^{w_{\mathrm{in}}-1}
(\bm{W}_u)_{o,c}\,(\bm{B}_u)_{i w_{\mathrm{out}}+j,\ i' w_{\mathrm{in}}+j'}\,A_{c,i',j'}.
\end{align}
The $(o,c)$ block of the Kronecker product $\bm{W}_u\otimes\bm{B}_u$ is $(\bm{W}_u)_{o,c}\bm{B}_u$. Thus, the last expression equals $[\sum_u(\bm{W}_u\otimes\bm{B}_u)\operatorname{vec}(\bm{A})]_{o n_{\mathrm{out}}+i w_{\mathrm{out}}+j}$. Collecting all output entries gives \cref{eq:conv-kronecker-draft}.
\end{proof}

Restoring the layer and model indices, \cref{prop:conv-matrix-draft} expresses, for example, convolution from layer $\ell$ to layer $\ell+1$ as
\begin{align}
\operatorname{vec}(\bm{Z}_{\ell+1}^{(m)})=\bm{T}_{\ell+1}^{(m)}\operatorname{vec}(\bm{A}_\ell^{(m)}),
\qquad
\bm{T}_{\ell+1}^{(m)}=\sum_{v\in\mathcal{K}_{\ell+1}}\bm{W}_{\ell+1,v}^{(m)}\otimes\bm{B}_{\ell+1,v}.
\label{eq:conv-layer-matrix-draft}
\end{align}
Below, we set $\bm{a}_\ell^{(m)}=\operatorname{vec}(\bm{A}_\ell^{(m)})$ to match the notation for activation vectors in the main text. Here, $d_\ell$ is the number of channels at layer $\ell$, and $n_\ell=h_\ell w_\ell$ is the number of spatial positions. Similarly, $\mathcal{K}_\ell$ is the set of scalar indices for the kernel positions at layer $\ell$. Models with the same architecture share the spatial matrices $\bm{B}_{\ell,u}$.

A channel permutation $\bm{P}_\ell$ reorders channels in the same way at every spatial position. Therefore, vectorizing in the order specified in \cref{eq:conv-vec-definition-draft} gives
\begin{align}
\operatorname{vec}(\widetilde{\bm{A}}_\ell^{(b)})
=(\bm{P}_\ell\otimes\bm{I}_{n_\ell})\operatorname{vec}(\bm{A}_\ell^{(b)}).
\end{align}
Write $\widehat{\bm{P}}_\ell=\bm{P}_\ell\otimes\bm{I}_{n_\ell}$ hereafter. Transforming each kernel matrix as $\widetilde{\bm{W}}_{\ell,u}^{(b)}=\bm{P}_\ell\bm{W}_{\ell,u}^{(b)}\bm{P}_{\ell-1}^\top$ to match the input and output channel permutations preserves the original network's input--output function. We search over the $d_\ell\times d_\ell$ matrix $\bm{P}_\ell$, so there is no need to search directly over a large permutation matrix that includes spatial positions.

\subsubsection{Objective for RO Weights}
\label{app:cnn_weights_objective}

We use the matrix representation above to define the RO Weights objective. Consider consecutive layers $\ell$ and $\ell+1$, and approximate the intervening ReLU by the identity, as in RO Weights in the main text. By \cref{eq:ro-weight-objective,eq:conv-layer-matrix-draft}, the sum of the squared norms of the cross terms is
\begin{align}
J_\ell^{\mathrm{conv}}(\bm{P}_\ell)
&=\|\bm{T}_{\ell+1}^{(a)}\widehat{\bm{P}}_\ell\bm{T}_\ell^{(b)}\|_F^2
+\|\bm{T}_{\ell+1}^{(b)}\widehat{\bm{P}}_\ell^\top\bm{T}_\ell^{(a)}\|_F^2.
\label{eq:ro-conv-draft}
\end{align}
Direct evaluation would require large convolution matrices. Expanding the Kronecker products separates the channel computations from the spatial information.

For this purpose, define the channel matrices for each pair of kernel positions $u\in\mathcal{K}_\ell$ and $v\in\mathcal{K}_{\ell+1}$ by
\begin{align}
\bm{M}_{vu}(\bm{P}_\ell)=\bm{W}_{\ell+1,v}^{(a)}\bm{P}_\ell\bm{W}_{\ell,u}^{(b)},
\qquad
\bm{N}_{vu}(\bm{P}_\ell)=\bm{W}_{\ell+1,v}^{(b)}\bm{P}_\ell^\top\bm{W}_{\ell,u}^{(a)}.
\end{align}
Both are $d_{\ell+1}\times d_{\ell-1}$ matrices.

Let $\bm{B}_{vu}^{\mathrm{path}}=\bm{B}_{\ell+1,v}\bm{B}_{\ell,u}$ be the corresponding spatial matrix, and define the scalar coefficient by the Frobenius inner product
\begin{align}
c_{(v,u),(v',u')}=\langle\bm{B}_{vu}^{\mathrm{path}},\bm{B}_{v'u'}^{\mathrm{path}}\rangle_F.
\label{eq:conv-spatial-gram-entry-draft}
\end{align}
This coefficient depends only on the spatial configuration of the convolutions, not on the weights or $\bm{P}_\ell$. These quantities give the following decomposition of the objective.

\begin{proposition}[Decomposition of the RO objective into channel matrices]
\label{prop:conv-ro-decomposition-draft}
Using the channel matrices and scalar coefficients above, the objective in \cref{eq:ro-conv-draft} can be written as
\begin{align}
J_\ell^{\mathrm{conv}}(\bm{P}_\ell)
&=\sum_{v,u,v',u'}c_{(v,u),(v',u')}\cdot\Bigl[
\langle\bm{M}_{vu}(\bm{P}_\ell),\bm{M}_{v'u'}(\bm{P}_\ell)\rangle_F
+\langle\bm{N}_{vu}(\bm{P}_\ell),\bm{N}_{v'u'}(\bm{P}_\ell)\rangle_F\Bigr].
\label{eq:conv-spatial-gram-draft}
\end{align}
The sum runs over $v,v'\in\mathcal{K}_{\ell+1}$ and $u,u'\in\mathcal{K}_\ell$. This expression is quadratic in the entries of $\bm{P}_\ell$, so minimizing it over permutation matrices is a QAP.
\end{proposition}
\begin{proof}
First, substitute \cref{eq:conv-layer-matrix-draft} into the first cross term. Using the Kronecker product identity $(\bm{A}\otimes\bm{B})(\bm{C}\otimes\bm{D})=(\bm{A}\bm{C})\otimes(\bm{B}\bm{D})$, we obtain
\begin{align}
\bm{T}_{\ell+1}^{(a)}\widehat{\bm{P}}_\ell\bm{T}_\ell^{(b)}
&=\left(\sum_v\bm{W}_{\ell+1,v}^{(a)}\otimes\bm{B}_{\ell+1,v}\right)
(\bm{P}_\ell\otimes\bm{I}_{n_\ell})
\left(\sum_u\bm{W}_{\ell,u}^{(b)}\otimes\bm{B}_{\ell,u}\right)
\nonumber\\
&=\sum_{v,u}\bigl(\bm{W}_{\ell+1,v}^{(a)}\bm{P}_\ell\bm{W}_{\ell,u}^{(b)}\bigr)
\otimes\bigl(\bm{B}_{\ell+1,v}\bm{B}_{\ell,u}\bigr)
\nonumber\\
&=\sum_{v,u}\bm{M}_{vu}(\bm{P}_\ell)\otimes\bm{B}_{vu}^{\mathrm{path}}.
\label{eq:conv-cross-kronecker-draft}
\end{align}
Next, expand the squared norm of this sum. The Frobenius inner product satisfies $\langle\bm{A}\otimes\bm{B},\bm{C}\otimes\bm{D}\rangle_F=\langle\bm{A},\bm{C}\rangle_F\langle\bm{B},\bm{D}\rangle_F$, so \cref{eq:conv-cross-kronecker-draft,eq:conv-spatial-gram-entry-draft} give
\begin{align}
\|\bm{T}_{\ell+1}^{(a)}\widehat{\bm{P}}_\ell\bm{T}_\ell^{(b)}\|_F^2
&=\sum_{v,u,v',u'}\left\langle
\bm{M}_{vu}(\bm{P}_\ell)\otimes\bm{B}_{vu}^{\mathrm{path}},
\bm{M}_{v'u'}(\bm{P}_\ell)\otimes\bm{B}_{v'u'}^{\mathrm{path}}\right\rangle_F
\nonumber\\
&=\sum_{v,u,v',u'}\langle\bm{M}_{vu}(\bm{P}_\ell),\bm{M}_{v'u'}(\bm{P}_\ell)\rangle_F
\,\langle\bm{B}_{vu}^{\mathrm{path}},\bm{B}_{v'u'}^{\mathrm{path}}\rangle_F
\nonumber\\
&=\sum_{v,u,v',u'}c_{(v,u),(v',u')}\cdot\langle\bm{M}_{vu}(\bm{P}_\ell),\bm{M}_{v'u'}(\bm{P}_\ell)\rangle_F.
\end{align}
The second term has the same expansion after exchanging models $a$ and $b$ and replacing $\bm{P}_\ell$ with $\bm{P}_\ell^\top$, with $\bm{N}_{vu}$ taking the place of $\bm{M}_{vu}$. Adding the two expressions gives \cref{eq:conv-spatial-gram-draft}. Since $\bm{M}_{vu}$ and $\bm{N}_{vu}$ are linear in $\bm{P}_\ell$, the sum of their inner products is quadratic.
\end{proof}

\Cref{prop:conv-ro-decomposition-draft} separates the evaluation of the objective into inner products of channel matrices and spatial coefficients $c_{(v,u),(v',u')}$. It remains to compute these coefficients without constructing large spatial matrices. The next proposition shows that these inner products can be obtained by counting output positions where two valid paths reach the same input location.

Here, a path $(v,u)$ traces locations backward from the output to the input, first through kernel position $v$ at layer $\ell+1$ and then through kernel position $u$ at layer $\ell$. Let $s_\ell,\delta_\ell,p_\ell$ be the stride, dilation, and padding at each layer, respectively, and let $\bm{\eta}_{\ell,u}=(r_{\ell,u},t_{\ell,u})^\top$ be the two-dimensional coordinates of a kernel position. Let $\Omega_{vu}$ be the set of output positions for which both the intermediate and input locations along path $(v,u)$ lie within their respective feature maps.

\begin{proposition}[Spatial coefficients as counts of valid paths]
\label{prop:conv-path-count-draft}
Let $\bm{d}_{vu}=\delta_\ell\bm{\eta}_{\ell,u}+s_\ell\delta_{\ell+1}\bm{\eta}_{\ell+1,v}$ be the combined offset of path $(v,u)$. The coefficient defined in \cref{eq:conv-spatial-gram-entry-draft} equals the number of output positions at which both paths are valid and access the same input position, and is given by
\begin{align}
c_{(v,u),(v',u')}
&=\begin{cases}
|\Omega_{vu}\cap\Omega_{v'u'}|, & \bm{d}_{vu}=\bm{d}_{v'u'},\\
0, & \bm{d}_{vu}\ne\bm{d}_{v'u'}
\end{cases}.
\label{eq:conv-path-count-draft}
\end{align}
Here, $|\cdot|$ denotes the cardinality of a set.
\end{proposition}
\begin{proof}
Let $\bm{q}=(i,j)^\top$ be an output position. Following path $(v,u)$ gives the intermediate location $\bm{q}_{\mathrm{mid}}=s_{\ell+1}\bm{q}-p_{\ell+1}\bm{1}+\delta_{\ell+1}\bm{\eta}_{\ell+1,v}$, where $\bm{1}=(1,1)^\top$. The final input location is therefore
\begin{align}
\bm{q}_{\mathrm{in}}
&=s_\ell\bm{q}_{\mathrm{mid}}-p_\ell\bm{1}+\delta_\ell\bm{\eta}_{\ell,u}
\nonumber\\
&=s_\ell s_{\ell+1}\bm{q}-(s_\ell p_{\ell+1}+p_\ell)\bm{1}+\bm{d}_{vu}.
\label{eq:conv-composed-offset-draft}
\end{align}
By \cref{eq:conv-composed-offset-draft}, two paths starting from the same output position access the same input position if and only if $\bm{d}_{vu}=\bm{d}_{v'u'}$.

Each row of $\bm{B}_{vu}^{\mathrm{path}}$ corresponds to one output position. If that position belongs to $\Omega_{vu}$, the row has a $1$ only in the column of the accessed input position. If the path accesses a padding region, the entire row is $0$. Thus, in the Frobenius inner product of two spatial matrices, a row contributes $1$ only when both paths are valid and access the same input position. No such row exists when the combined offsets differ. When they agree, each row corresponding to $\Omega_{vu}\cap\Omega_{v'u'}$ contributes $1$. This establishes \cref{eq:conv-path-count-draft}.
\end{proof}

This proposition identifies $c_{(v,u),(v',u')}$ as a nonnegative integer counting how often the two paths connect the same input and output positions. In particular, if all strides and dilations are $1$, the combined offsets agree exactly when $\bm{\eta}_{\ell,u}+\bm{\eta}_{\ell+1,v}=\bm{\eta}_{\ell,u'}+\bm{\eta}_{\ell+1,v'}$. If neither layer uses padding, every path is valid at every output position, so the coefficient equals the number of output positions $n_{\ell+1}$ when the combined offsets agree. With padding, paths reaching the same input position can pass through different intermediate positions and therefore have different sets of valid output positions. We must therefore count the valid positions common to both paths.

Thus, the RO objective can be evaluated using path counts and inner products of channel matrices without constructing large convolution or spatial matrices.

When evaluating RO Weights, we replace ReLU with the identity and omit biases and BatchNorm. We also exclude max pooling between VGG convolutional layers from the product of the two layer matrices above. When pooling lies between the layers, the resulting objective is therefore an approximation based on convolution matrices constructed on a grid that omits pooling.

\subsubsection{Objective for RO Features}
\label{app:cnn_features_objective}

RO Features evaluates the cross terms in \cref{sec:ro-activation} using the feature maps actually supplied to each convolutional layer during forward passes of the original models. Let $\bm{H}_\ell^{(m)}(\bm{x})\in\mathbb{R}^{d_\ell\times n_\ell}$ be the matrix obtained by rearranging the input feature map $\bm{A}_\ell^{(m)}(\bm{x})$ to layer $\ell+1$, with channels as rows and spatial positions as columns. Specifically, $(\bm{H}_\ell^{(m)}(\bm{x}))_{c,i w_\ell+j}=(\bm{A}_\ell^{(m)}(\bm{x}))_{c,i,j}$. Using the spatial selection matrix defined above, define the patch matrix for kernel position $u\in\mathcal{K}_{\ell+1}$ by
\begin{align}
\bm{X}_{\ell,u}^{(m)}(\bm{x})=\bm{H}_\ell^{(m)}(\bm{x})\bm{B}_{\ell+1,u}^\top
\in\mathbb{R}^{d_\ell\times n_{\ell+1}}.
\label{eq:conv-feature-patch-draft}
\end{align}
Each column of this matrix is the input channel vector accessed by kernel position $u$ at a particular output position. Thus, $\sum_u\bm{W}_{\ell+1,u}^{(m)}\bm{X}_{\ell,u}^{(m)}(\bm{x})$ represents the convolution output without bias as a matrix indexed by channels and spatial positions.

Permuting the input channels of model $b$ by $\bm{P}_\ell$ and applying the kernel of model $a$ gives the cross term $\sum_u\bm{W}_{\ell+1,u}^{(a)}\bm{P}_\ell\bm{X}_{\ell,u}^{(b)}(\bm{x})$. Including the reverse cross term, we minimize the following objective computed from the $N$ images $\{\bm{x}_i\}_{i=1}^N$ used for feature extraction
\begin{align}
J_\ell^{\mathrm{feat}}(\bm{P}_\ell)
&=\frac{1}{N}\sum_{i=1}^N\left[
\left\|\sum_u\bm{W}_{\ell+1,u}^{(a)}\bm{P}_\ell\bm{X}_{\ell,u}^{(b)}(\bm{x}_i)\right\|_F^2
+\left\|\sum_u\bm{W}_{\ell+1,u}^{(b)}\bm{P}_\ell^\top\bm{X}_{\ell,u}^{(a)}(\bm{x}_i)\right\|_F^2
\right].
\label{eq:conv-feature-objective-draft}
\end{align}
Each Frobenius norm sums over output channels and spatial positions, and we average only over the number of images $N$. All sums over kernel positions here run over $u\in\mathcal{K}_{\ell+1}$.

To express this objective as a quadratic form, define the uncentered second moment for each pair of kernel positions by
\begin{align}
\bm{C}_{\ell,uv}^{(m)}=\frac{1}{N}\sum_{i=1}^N
\bm{X}_{\ell,u}^{(m)}(\bm{x}_i)\bigl(\bm{X}_{\ell,v}^{(m)}(\bm{x}_i)\bigr)^\top
\in\mathbb{R}^{d_\ell\times d_\ell}.
\label{eq:conv-feature-moment-draft}
\end{align}
In the notation of $\bm{G}_{\mathrm{col},\ell+1}^{(m)}$ in the main text, define the column inner products for each pair of kernel positions by
\begin{align}
\bm{G}_{\mathrm{col},\ell+1,vu}^{(m)}
=(\bm{W}_{\ell+1,v}^{(m)})^\top\bm{W}_{\ell+1,u}^{(m)}
\in\mathbb{R}^{d_\ell\times d_\ell}.
\label{eq:conv-column-products-draft}
\end{align}
When $v=u$, this is the column Gram matrix of a single kernel matrix. When $v\ne u$, it contains inner products between columns at different kernel positions and is generally not symmetric. Together with the second moments, these matrices give the following quadratic form.

\begin{proposition}[Quadratic form of RO Features]
\label{prop:conv-feature-quadratic-draft}
Using the second moments in \cref{eq:conv-feature-moment-draft}, the objective in \cref{eq:conv-feature-objective-draft} can be written as
\begin{align}
J_\ell^{\mathrm{feat}}(\bm{P}_\ell)
&=\sum_{u,v\in\mathcal{K}_{\ell+1}}\left[
\langle\bm{P}_\ell,\bm{G}_{\mathrm{col},\ell+1,vu}^{(a)}\bm{P}_\ell\bm{C}_{\ell,uv}^{(b)}\rangle_F
+\langle\bm{P}_\ell,\bm{C}_{\ell,uv}^{(a)}\bm{P}_\ell\bm{G}_{\mathrm{col},\ell+1,vu}^{(b)}\rangle_F
\right].
\label{eq:conv-feature-quadratic-draft}
\end{align}
Thus, searching for the channel permutation $\bm{P}_\ell$ is a QAP of the same form as in RO Weights.
\end{proposition}
\begin{proof}
Expanding the squared norm of the first cross term and using the cyclic property of the trace and \cref{eq:conv-feature-moment-draft} gives
\begin{align}
\frac{1}{N}\sum_{i=1}^N\left\|\sum_u\bm{W}_{\ell+1,u}^{(a)}\bm{P}_\ell\bm{X}_{\ell,u}^{(b)}(\bm{x}_i)\right\|_F^2
&=\frac{1}{N}\sum_{i=1}^N\sum_{u,v}\operatorname{tr}\!\left(
\bm{P}_\ell^\top\bm{G}_{\mathrm{col},\ell+1,vu}^{(a)}\bm{P}_\ell
\bm{X}_{\ell,u}^{(b)}(\bm{x}_i)\bigl(\bm{X}_{\ell,v}^{(b)}(\bm{x}_i)\bigr)^\top\right)
\nonumber\\
&=\sum_{u,v}\operatorname{tr}\!\left(\bm{P}_\ell^\top\bm{G}_{\mathrm{col},\ell+1,vu}^{(a)}\bm{P}_\ell\bm{C}_{\ell,uv}^{(b)}\right)
\nonumber\\
&=\sum_{u,v}\langle\bm{P}_\ell,\bm{G}_{\mathrm{col},\ell+1,vu}^{(a)}\bm{P}_\ell\bm{C}_{\ell,uv}^{(b)}\rangle_F.
\end{align}
Expanding the second term in the same way gives $\sum_{u,v}\operatorname{tr}(\bm{P}_\ell\bm{G}_{\mathrm{col},\ell+1,vu}^{(b)}\bm{P}_\ell^\top\bm{C}_{\ell,uv}^{(a)})$. Applying the cyclic property of the trace again gives $\sum_{u,v}\langle\bm{P}_\ell,\bm{C}_{\ell,uv}^{(a)}\bm{P}_\ell\bm{G}_{\mathrm{col},\ell+1,vu}^{(b)}\rangle_F$. Adding the two expressions yields \cref{eq:conv-feature-quadratic-draft}.
\end{proof}


\subsubsection{Computational Procedure for Permutation Search}
\label{app:cnn_computation}

As in the main text, we precompute inner products of weight columns and rows and second moments of features, then hold these quantities fixed during permutation search. The relaxed objectives of both RO Weights and RO Features can be written using fixed matrices $\bm{L}_q,\bm{R}_q\in\mathbb{R}^{d_\ell\times d_\ell}$ as
\begin{align}
J_\ell(\bm{X})=\sum_q\langle\bm{X},\bm{L}_q\bm{X}\bm{R}_q\rangle_F.
\label{eq:conv-factorized-objective-draft}
\end{align}
Here, $J_\ell$ is either $J_\ell^{\mathrm{conv}}$ or $J_\ell^{\mathrm{feat}}$, and $\bm{X}$ is a doubly stochastic relaxation of the permutation matrix. The matrices $\bm{L}_q$ and $\bm{R}_q$ multiply $\bm{X}$ from the left and right in each term, and $q$ indexes the combinations of kernel positions and the direction of the cross term. We give the specific construction for each method below.

For RO Weights, the decomposition in \cref{prop:conv-ro-decomposition-draft} gives the following procedure.
\begin{enumerate}
\item Precompute the coefficients $c_{(v,u),(v',u')}$ in \cref{eq:conv-path-count-draft} from the spatial configuration of the convolutions. The coefficient is $0$ when the combined offsets differ. When they agree, count the positions at which both paths are valid separately along the height and width dimensions and multiply the two counts. Thus, there is no need to store $\bm{B}_{\ell,u}$ as a large matrix.
\item In addition to the column inner products in \cref{eq:conv-column-products-draft}, define the row inner products corresponding to $\bm{G}_{\mathrm{row},\ell}^{(m)}$ in the main text by
\begin{align}
\bm{G}_{\mathrm{row},\ell,u'u}^{(m)}
=\bm{W}_{\ell,u'}^{(m)}(\bm{W}_{\ell,u}^{(m)})^\top
\in\mathbb{R}^{d_\ell\times d_\ell}.
\label{eq:conv-row-products-draft}
\end{align}
Precompute these matrices for each pair of kernel positions. When $u'=u$, this is the row Gram matrix of a single kernel matrix. By the cyclic property of the trace, each term of the objective can be written as
\begin{align}
\langle\bm{M}_{vu}(\bm{P}_\ell),\bm{M}_{v'u'}(\bm{P}_\ell)\rangle_F
&=\operatorname{tr}\bigl(\bm{P}_\ell^\top\bm{G}_{\mathrm{col},\ell+1,vv'}^{(a)}\bm{P}_\ell\bm{G}_{\mathrm{row},\ell,u'u}^{(b)}\bigr)
\nonumber\\
&=\langle\bm{P}_\ell,\bm{G}_{\mathrm{col},\ell+1,vv'}^{(a)}\bm{P}_\ell\bm{G}_{\mathrm{row},\ell,u'u}^{(b)}\rangle_F,
\nonumber\\
\langle\bm{N}_{vu}(\bm{P}_\ell),\bm{N}_{v'u'}(\bm{P}_\ell)\rangle_F
&=\langle\bm{P}_\ell,\bm{G}_{\mathrm{row},\ell,u'u}^{(a)}\bm{P}_\ell\bm{G}_{\mathrm{col},\ell+1,vv'}^{(b)}\rangle_F.
\label{eq:conv-channel-quadratic-terms-draft}
\end{align}
Weighting the two terms in \cref{eq:conv-channel-quadratic-terms-draft} by $c_{(v,u),(v',u')}$ and summing them gives a quadratic expression corresponding to \cref{eq:ro-weight-objective} in the main text. Specifically, let $q=(v,u,v',u',t)$ be the index, with $v,v'\in\mathcal{K}_{\ell+1}$, $u,u'\in\mathcal{K}_\ell$, and $t\in\{1,2\}$, and set
\begin{align}
(\bm{L}_q,\bm{R}_q)
&=\begin{cases}
\bigl(c_{(v,u),(v',u')}\bm{G}_{\mathrm{col},\ell+1,vv'}^{(a)},\ \bm{G}_{\mathrm{row},\ell,u'u}^{(b)}\bigr), & t=1,\\
\bigl(c_{(v,u),(v',u')}\bm{G}_{\mathrm{row},\ell,u'u}^{(a)},\ \bm{G}_{\mathrm{col},\ell+1,vv'}^{(b)}\bigr), & t=2
\end{cases}.
\label{eq:conv-weight-factors-draft}
\end{align}
These choices give \cref{eq:conv-factorized-objective-draft}. The cases $t=1$ and $t=2$ correspond to the terms from $\bm{M}$ and $\bm{N}$, respectively. Here, we include the spatial coefficient in $\bm{L}_q$. Our method groups terms in \cref{eq:conv-weight-factors-draft} that share the same pattern of spatial coefficients before the search to reduce the computational cost.
\item Evaluate the objective and gradient using the fixed matrices. Reversing the pair of kernel positions in \cref{eq:conv-column-products-draft,eq:conv-row-products-draft} transposes each matrix, and the coefficients satisfy $c_{(v,u),(v',u')}=c_{(v',u'),(v,u)}$. Thus, even when individual $\bm{L}_q,\bm{R}_q$ are not symmetric, summing all terms gives
\begin{align}
\nabla J_\ell(\bm{X})=2\sum_q\bm{L}_q\bm{X}\bm{R}_q,
\qquad
J_\ell(\bm{X})=\tfrac12\langle\bm{X},\nabla J_\ell(\bm{X})\rangle_F.
\label{eq:conv-factorized-gradient-draft}
\end{align}
These expressions correspond to \cref{eq:ro-gradient,eq:ro-objective-gradient} in the main text. As in \cref{sec:ro-frank-wolfe}, we repeatedly solve a linear assignment problem using the gradient, perform line search, and update $\bm{X}$, then select a permutation matrix at the end. Only $\bm{X}$, which represents the channel correspondence, changes during the iterations, so the spatial quantities do not need to be recomputed.
\end{enumerate}

In our PyTorch implementation of RO Features, we run the original models in evaluation mode and extract the patches corresponding to \cref{eq:conv-feature-patch-draft}. We stack $\bm{X}_{\ell,u}^{(m)}(\bm{x}_i)$ vertically in the order of kernel positions and concatenate the resulting matrices horizontally across images. For each batch, we accumulate the product of this matrix and its transpose. Dividing by $N$ at the end gives a matrix whose blocks are $\bm{C}_{\ell,uv}^{(m)}$. We retain the terms with $u\ne v$, evaluating the objective that sums contributions from different kernel positions before squaring. Since the second moments are precomputed, permutation search does not require repeated forward passes over the images.

For RO Features, let $q=(u,v,t)$ be the index, with $u,v\in\mathcal{K}_{\ell+1}$ and $t\in\{1,2\}$. By \cref{eq:conv-feature-quadratic-draft}, set
\begin{align}
(\bm{L}_q,\bm{R}_q)
&=\begin{cases}
\bigl(\bm{G}_{\mathrm{col},\ell+1,vu}^{(a)},\ \bm{C}_{\ell,uv}^{(b)}\bigr), & t=1,\\
\bigl(\bm{C}_{\ell,uv}^{(a)},\ \bm{G}_{\mathrm{col},\ell+1,vu}^{(b)}\bigr), & t=2
\end{cases}.
\label{eq:conv-feature-factors-draft}
\end{align}
The case $t=1$ corresponds to applying the weights of model $a$ to the features of model $b$, and $t=2$ corresponds to the reverse cross term. In \cref{eq:conv-feature-factors-draft}, the identities $\bm{C}_{\ell,vu}^{(m)}=(\bm{C}_{\ell,uv}^{(m)})^\top$ and $\bm{G}_{\mathrm{col},\ell+1,uv}^{(m)}=(\bm{G}_{\mathrm{col},\ell+1,vu}^{(m)})^\top$ give the same gradient formula in \cref{eq:conv-factorized-gradient-draft}. We hold these matrices fixed and search for permutations using the same Frank--Wolfe procedure as for RO Weights.

\subsection{RO and LEWC with Residual Connections}
\label{app:resnet_lewc}

This section explains how merging by parameter addition behaves under RO in architectures with skip connections, such as ResNet-20. Specifically, assuming RO and ReLU weak additivity, we show that each merged residual block preserves the sum of the original models' outputs, enabling ensemble-like computation throughout the network.

Consider $\bm{\theta}^{(c)}=\bm{\theta}^{(a)}+\pi(\bm{\theta}^{(b)})$, obtained by adding the parameters of models $a$ and $b$. Below, the weights and features of model $b$ refer to those after permutation, and we omit biases and BatchNorm. We assume that each trainable operation acting on intermediate features satisfies RO with respect to the original models' features, and that each ReLU satisfies weak additivity. Multiplying the input to ReLU by a nonnegative constant multiplies its output by the same constant. Thus, weak additivity in \cref{def:weak-additivity} also applies to sums with nonnegative coefficients that do not necessarily sum to one. Parameter addition adds the trainable weights, while an identity skip connection remains the identity.

First, consider a single residual block. For model $m$ ($m\in\{a,b\}$), let $\bm{h}^{(m)}$ be the block input, and let $\bm{T}_1^{(m)}$ and $\bm{T}_2^{(m)}$ be the two convolution matrices in the main branch. Write the output of the main branch as $\bm{u}^{(m)}=\bm{T}_2^{(m)}\sigma(\bm{T}_1^{(m)}\bm{h}^{(m)})$. Assume that the input to the merged block satisfies $\bm{h}^{(c)}=\bm{h}^{(a)}+\bm{h}^{(b)}$. We show that this relation also holds at the block output by considering the main branch and then the skip branch.

In the main branch, parameter addition gives $\bm{T}_1^{(c)}=\bm{T}_1^{(a)}+\bm{T}_1^{(b)}$. The output of the first convolution is therefore
\begin{align}
\bm{T}_1^{(c)}\bm{h}^{(c)}
&=\bm{T}_1^{(a)}\bm{h}^{(a)}+\bm{T}_1^{(a)}\bm{h}^{(b)}
+\bm{T}_1^{(b)}\bm{h}^{(a)}+\bm{T}_1^{(b)}\bm{h}^{(b)}
\nonumber\\
&=\bm{T}_1^{(a)}\bm{h}^{(a)}+\bm{T}_1^{(b)}\bm{h}^{(b)}.
\label{eq:resnet-sum-cross-terms-draft}
\end{align}
The last equality follows from RO, which makes the cross terms $\bm{T}_1^{(a)}\bm{h}^{(b)}$ and $\bm{T}_1^{(b)}\bm{h}^{(a)}$ equal to $\bm{0}$. Applying ReLU weak additivity to \cref{eq:resnet-sum-cross-terms-draft} shows that the activation after the first convolution is also the sum of the original models' activations. At the second convolution, $\bm{T}_2^{(c)}=\bm{T}_2^{(a)}+\bm{T}_2^{(b)}$ and RO again eliminate the cross terms, giving the output of the main branch $\bm{u}^{(c)}=\bm{u}^{(a)}+\bm{u}^{(b)}$.

Next, consider the identity skip connection. This branch passes the block input $\bm{h}^{(c)}=\bm{h}^{(a)}+\bm{h}^{(b)}$ through unchanged. Adding the main and skip branches therefore gives $(\bm{u}^{(a)}+\bm{h}^{(a)})+(\bm{u}^{(b)}+\bm{h}^{(b)})$. Let $\bm{h}_{\mathrm{out}}^{(m)}=\sigma(\bm{u}^{(m)}+\bm{h}^{(m)})$ be the block output of model $m$. Weak additivity of the ReLU after residual addition then gives
\begin{align}
\bm{h}_{\mathrm{out}}^{(c)}
&=\sigma\!\left((\bm{u}^{(a)}+\bm{h}^{(a)})+(\bm{u}^{(b)}+\bm{h}^{(b)})\right)
=\bm{h}_{\mathrm{out}}^{(a)}+\bm{h}_{\mathrm{out}}^{(b)}.
\end{align}
Thus, the sum relation at the input is preserved at the block output.

The same argument applies to a trainable skip branch. Let $\bm{S}^{(m)}$ be its convolution matrix. Parameter addition gives $\bm{S}^{(c)}=\bm{S}^{(a)}+\bm{S}^{(b)}$. If RO gives $\bm{S}^{(a)}\bm{h}^{(b)}=\bm{S}^{(b)}\bm{h}^{(a)}=\bm{0}$, this branch outputs $\bm{S}^{(a)}\bm{h}^{(a)}+\bm{S}^{(b)}\bm{h}^{(b)}$. Adding the output of the main branch and applying weak additivity to the ReLU after residual addition again gives the sum of the original models' block outputs.

Finally, apply this relation throughout the network. At the input layer, the sum of the convolution outputs for the shared input and ReLU weak additivity give the sum of the original models' activations. The argument above then propagates this sum relation through the subsequent residual blocks. If the final pooling operation is linear and the classifier also satisfies RO, the logits are the sum of the original models' logits. Under these assumptions, parameter addition therefore allows the original models' computations to proceed in parallel and their outputs to be added, even in the presence of skip connections.



\FloatBarrier
\section{Additional Experimental Results}
\label{app:additional_results}

\subsection{Geometry of the Low-Loss Region}
\label{app:loss_landscape}

\begin{figure}[tbp]
\centering
\includegraphics[width=\textwidth]{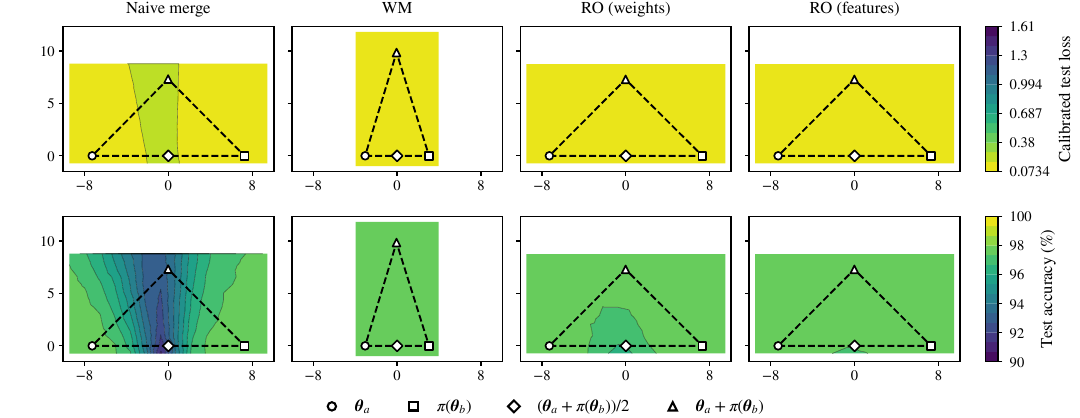}
\caption{Test loss after softmax temperature calibration (top) and test accuracy (bottom) for a single pair of MLPs on MNIST in the parameter-space plane. Panel and marker conventions follow \cref{fig:landscape-cifar10-vgg}.}
\label{fig:landscape-mnist-mlp}
\end{figure}

\begin{figure}[tbp]
\centering
\includegraphics[width=\textwidth]{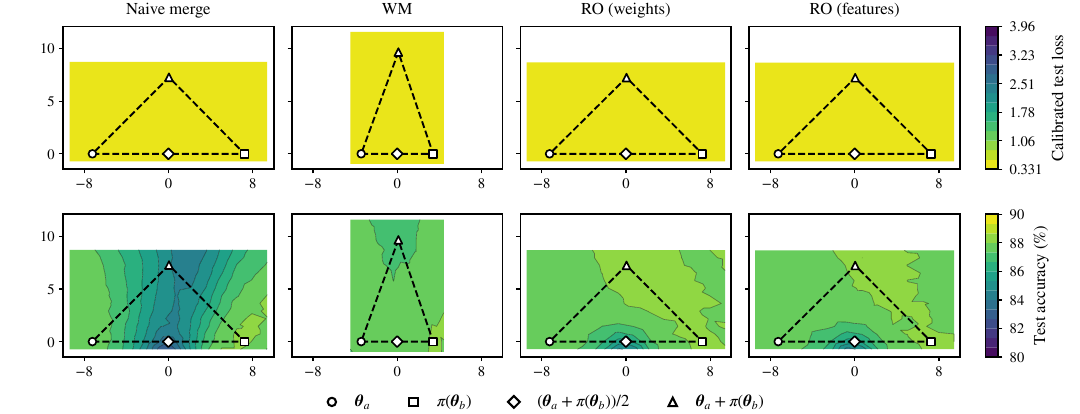}
\caption{Test loss after softmax temperature calibration (top) and test accuracy (bottom) for a single pair of MLPs on Fashion-MNIST in the parameter-space plane. Panel and marker conventions follow \cref{fig:landscape-cifar10-vgg}.}
\label{fig:landscape-fmnist-mlp}
\end{figure}

\begin{figure}[tbp]
\centering
\includegraphics[width=\textwidth]{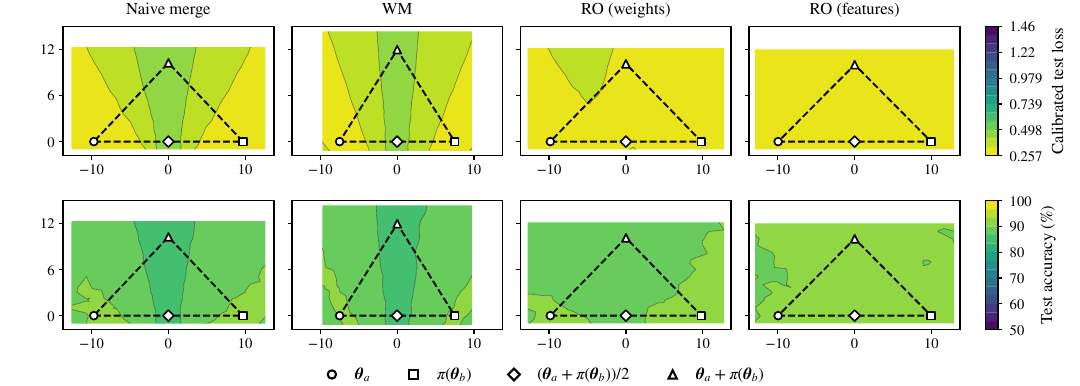}
\caption{Test loss after softmax temperature calibration (top) and test accuracy (bottom) for a single pair of ResNet-20 models on CIFAR-10 in the parameter-space plane. Panel and marker conventions follow \cref{fig:landscape-cifar10-vgg}.}
\label{fig:landscape-cifar10-resnet20}
\end{figure}

\begin{figure}[tbp]
\centering
\includegraphics[width=\textwidth]{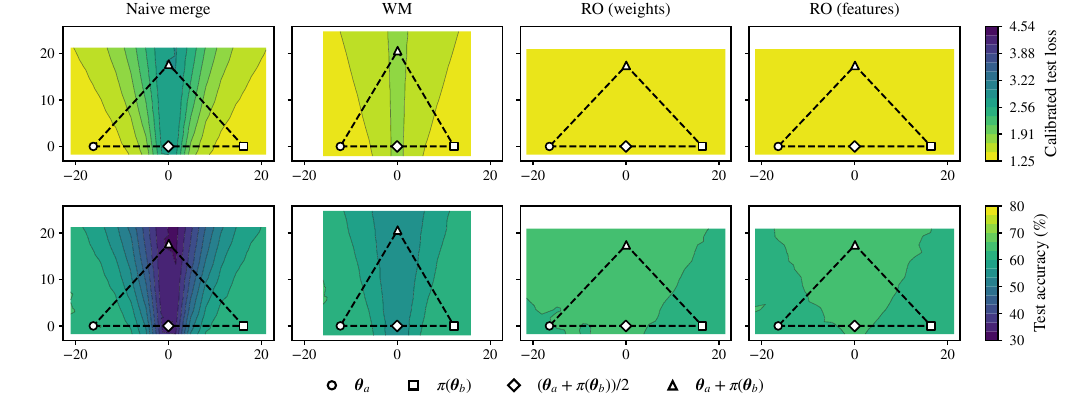}
\caption{Test loss after softmax temperature calibration (top) and test accuracy (bottom) for a single pair of VGG-11 models on CIFAR-100 in the parameter-space plane. Panel and marker conventions follow \cref{fig:landscape-cifar10-vgg}.}
\label{fig:landscape-cifar100-vgg}
\end{figure}

\begin{figure}[tbp]
\centering
\includegraphics[width=\textwidth]{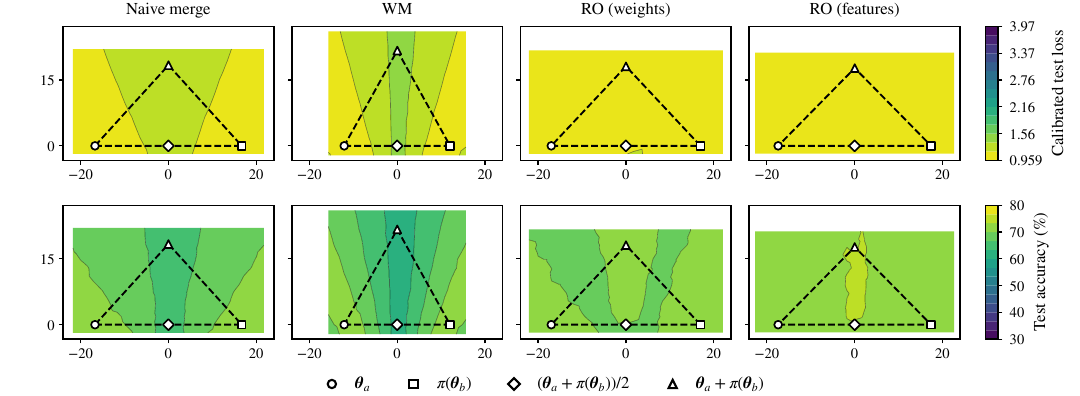}
\caption{Test loss after softmax temperature calibration (top) and test accuracy (bottom) for a single pair of ResNet-20 models on CIFAR-100 in the parameter-space plane. Panel and marker conventions follow \cref{fig:landscape-cifar10-vgg}.}
\label{fig:landscape-cifar100-resnet20}
\end{figure}

\Cref{fig:landscape-mnist-mlp,fig:landscape-fmnist-mlp,fig:landscape-cifar10-resnet20,fig:landscape-cifar100-vgg,fig:landscape-cifar100-resnet20} supplement the VGG-11 results on CIFAR-10 in \cref{sec:exp-loss-landscape} with the five remaining dataset--architecture combinations. Each figure shows test loss after softmax temperature calibration in the top row and test accuracy in the bottom row. Columns and markers follow the conventions of \cref{fig:landscape-cifar10-vgg}.

\clearpage
\subsection{Effect of Network Width}
\label{app:width_dependence}

\begin{figure}[tbp]
\centering
\includegraphics[width=\textwidth]{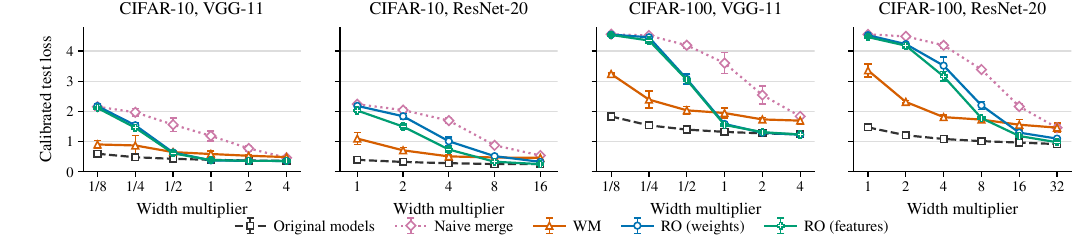}
\caption{Effect of network width on test loss after softmax temperature calibration for original and merged models. Results are shown for VGG-11 and ResNet-20 on CIFAR-10 and CIFAR-100, corresponding to the test-accuracy results in \cref{fig:width-comparison}. Error bars indicate sample standard deviations over three model pairs.}
\label{fig:width-calibrated-test-loss}
\end{figure}

\Cref{fig:width-calibrated-test-loss} shows test loss after softmax temperature calibration for the network-width comparison in \cref{sec:exp-same-conditions}. The corresponding test accuracies are shown in \cref{fig:width-comparison}. The compared models and methods are the same in both figures.

\end{document}